\documentclass[11pt]{article}

\usepackage[final]{acl}

\usepackage{times}
\usepackage{latexsym}

\usepackage[T1]{fontenc}

\usepackage[utf8]{inputenc}

\usepackage{microtype}

\usepackage{inconsolata}

\usepackage{graphicx}

\title{SERE: Structural Example Retrieval for Enhancing LLMs in Event Causality Identification}

\author{
 \textbf{Zhifeng Hao\textsuperscript{1,2}},
 \textbf{Zhongjie Chen\textsuperscript{1}},
 \textbf{Junhao Lu\textsuperscript{1}},
 \textbf{Shengyin Yu\textsuperscript{1}} \\
 \textbf{Guimin Hu\textsuperscript{1}}, 
 \textbf{Keli Zhang\textsuperscript{3}},
 \textbf{Ruichu Cai\textsuperscript{1,4}},
 \textbf{Boyan Xu\textsuperscript{1}}\thanks{Corresponding author.}
 \\
 \textsuperscript{1}School of Computer Science, Guangdong University of Technology \\
 \textsuperscript{2}College of Mathematics and Computer, Shantou University \\
  \textsuperscript{3}Huawei Noah's Ark Lab\quad \textsuperscript{4}Peng Cheng Laboratory \\
 haozhifeng@stu.edu.cn \quad 
 \{bariancgg, lujunhao.code, yushengyin2022\}@gmail.com \\
 rice.hu.x@gmail.com \quad zhangkeli1@huawei.com \quad
  \{cairuichu, hpakyim\}@gmail.com
}

\usepackage{amssymb}
\usepackage{algorithm}
\usepackage{algpseudocode}
\usepackage{amsmath}
\usepackage{float}
\usepackage{booktabs}
\usepackage{multirow}
\usepackage{tikz}
\usepackage{graphicx}
\usepackage{caption}
\usepackage{listings}
\usepackage{xcolor}
\usepackage[most]{tcolorbox}

\lstdefinelanguage{Cypher}{
  morekeywords={
    MATCH, RETURN, WHERE, AND, OR, NOT, IN, AS, 
    CREATE, DELETE, DETACH, REMOVE, SET, MERGE, 
    ON, OPTIONAL, WITH, DISTINCT, LIMIT, ORDER, BY,
    UNION, ALL, UNWIND, FOREACH, CASE, WHEN, THEN, ELSE, END,
    shortestPath, relationships, startNode, endNode, type
  },
  sensitive=true,
  morecomment=[l]{//},
  morestring=[b]",
}
\lstdefinestyle{neo4j}{
  language=Cypher,
  basicstyle=\ttfamily\fontsize{8pt}{15pt}\selectfont,
  keywordstyle=\color{blue!80!black}\bfseries,
  commentstyle=\color{gray},
  stringstyle=\color{green!60!black},
  showstringspaces=false,
  breaklines=true,
  frame=single,
  columns=fullflexible,
  captionpos=b
}
\lstdefinelanguage{PythonCustom}{
  morekeywords={
    and, as, assert, break, class, continue, def, del, elif, else, 
    except, False, finally, for, from, global, if, import, in, 
    is, lambda, None, nonlocal, not, or, pass, raise, return, 
    True, try, while, with, yield
  },
  sensitive=true,
  morecomment=[l]{\#},
  morestring=[b]',
  morestring=[b]",
}
\lstdefinestyle{pythonstyle}{
  language=PythonCustom,
  basicstyle=\ttfamily\fontsize{8pt}{15pt}\selectfont,
  keywordstyle=\color{blue!80!black}\bfseries,
  commentstyle=\color{gray},
  stringstyle=\color{green!60!black},
  showstringspaces=false,
  breaklines=true,
  frame=single,
  columns=fullflexible,
  captionpos=b
}

\newtcolorbox[auto counter]{prompt-box}[2][]{%
  colbacktitle=gray!50,      %
  colframe=black!60,
  colback=white,
  width=\columnwidth,
  arc=2mm, auto outer arc,
  title={Prompt \thetcbcounter: #2},
  fonttitle=\Large,
  fontupper=\footnotesize,
  before upper=\obeylines,
  #1
}

\newcommand\eg{\emph{e.g.}}

\newtcolorbox[auto counter]{case-box}[2][]{%
  colbacktitle=gray!50,      %
  colframe=black!60,
  colback=white,
  width=\textwidth,
  arc=2mm, auto outer arc,
  title={Case \thetcbcounter: #2},
  fonttitle=\Large,
  fontupper=\footnotesize,
  before upper=\obeylines,
  #1
}

\begin{document}
\maketitle
\begin{abstract}
Event Causality Identification (ECI) requires models to determine whether a given pair of events in a context exhibits a causal relationship. While Large Language Models (LLMs) have demonstrated strong performance across various NLP tasks, their effectiveness in ECI remains limited due to biases in causal reasoning, often leading to overprediction of causal relationships (causal hallucination). To mitigate these issues and enhance LLM performance in ECI, we propose \textbf{SERE}, a structural example retrieval framework that leverages LLMs’ few-shot learning capabilities. \textbf{SERE} introduces an innovative retrieval mechanism based on three structural concepts: (i) \textbf{Conceptual Path Metric}, which measures the conceptual relationship between events using edit distance in ConceptNet; (ii) \textbf{Syntactic Metric}, which quantifies structural similarity through tree edit distance on syntactic trees; and (iii) \textbf{Causal Pattern Filtering}, which filters examples based on predefined causal structures using LLMs. By integrating these structural retrieval strategies, \textbf{SERE} selects more relevant examples to guide LLMs in causal reasoning, mitigating bias and improving accuracy in ECI tasks. Extensive experiments on multiple ECI datasets validate the effectiveness of \textbf{SERE}. The source code is publicly available at \url{https://github.com/DMIRLAB-Group/SERE}.
\end{abstract}

\section{Introduction}

\begin{figure}[!h]
    \includegraphics[width=\columnwidth]{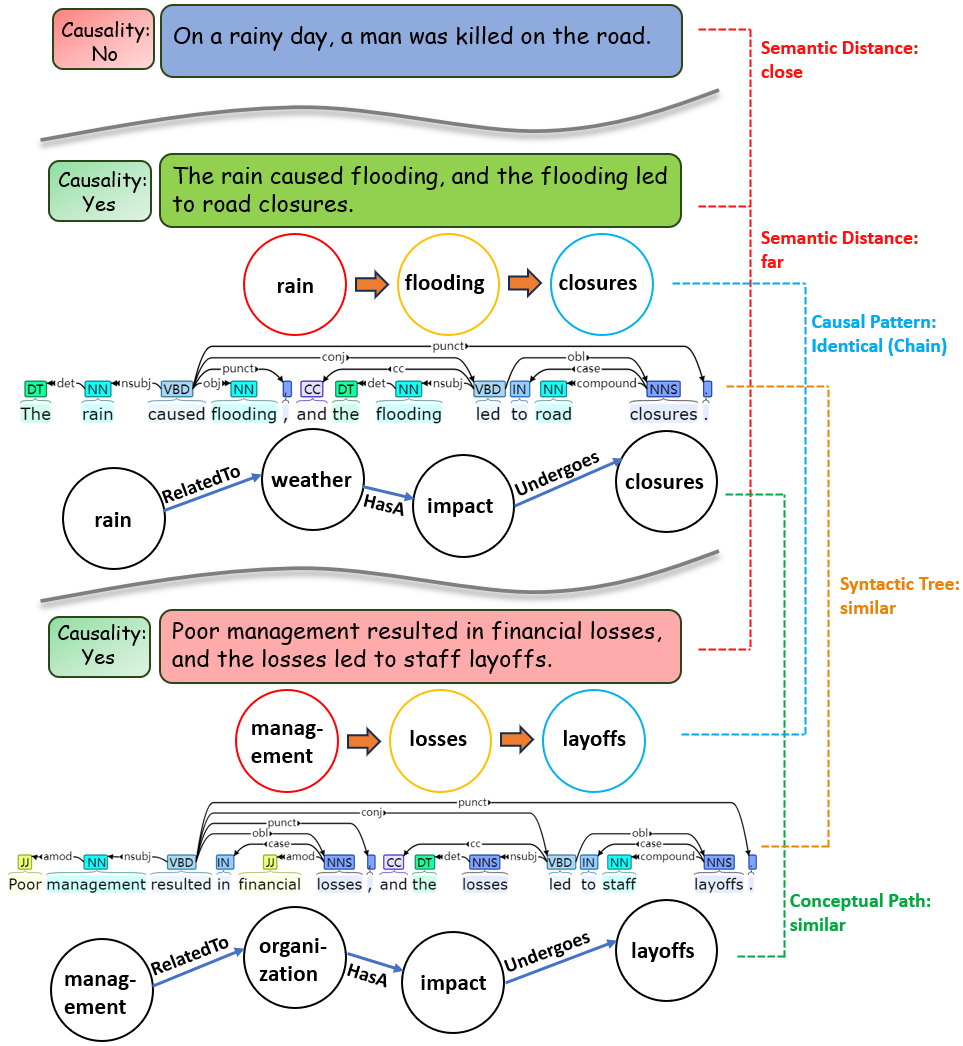}
    \caption{The middle instance is the sample to be inferred, while the two instances above and below are the corpus samples to be retrieved. The upper sample has a similar semantic similarity to the middle one but differs in causality. Selecting this sample may mislead the LLM. The lower sample has low semantic similarity to the sample to be inferred, but it is structurally similar and shares the same causality. Selecting this example helps guide the LLM to reason correctly.}
    \label{fig:example}
\end{figure}

Event Causality Identification (ECI) requires models to determine whether a given pair of events within a context exhibits a causal relationship \cite{zuo-etal-2020-knowdis, dunietz-etal-2015-annotating}. Traditionally, ECI methods have relied on fine-tuning pre-trained language models, primarily encoder-only architectures such as BERT \cite{devlin2019bertpretrainingdeepbidirectional} and RoBERTa \cite{liu2019robertarobustlyoptimizedbert}. However, Although fine-tuning enables the model to achieve higher performance on the ECI task, due to the limited size of available ECI datasets, fine-tuned models often struggle to generalize effectively. With the emergence of Large Language Models (LLMs) such as GPT \cite{Radford2018ImprovingLU}, which leverage vast amounts of corpus data, it has become possible to apply these models to ECI without requiring additional training. This enables fast deployment of LLMs in ECI tasks.
However, recent studies \cite{gao2023chatgptgoodcausalreasoner} have revealed that applying LLMs to ECI may introduce biases in causal reasoning, leading to a tendency to overpredict causal relationships—a phenomenon referred to as causal hallucination. This raises a fundamental challenge in adapting LLMs for accurate ECI.

To mitigate bias and improve the model's ability to capture complex causal relationships, few-shot learning has emerged as a promising approach, where the effectiveness heavily depends on the example selection strategy \cite{liu2021makesgoodincontextexamples, dong2024surveyincontextlearning}. However, existing methods primarily rely on semantic retrieval, which presents notable limitations: 
In ECI, causal relationships between events are inherently structural rather than purely semantic. For instance, as illustrated in Figure~\ref{fig:example}, semantic retrieval methods may select examples based on shared words like ``rain'' and ``road'' appearing in both the first and second instances. However, these two instances exhibit opposite causality. Relying solely on semantic similarity can therefore lead to incorrect example selection, potentially causing LLMs to misidentify causal relationships.

To retrieve more structurally relevant examples and address the limitations of semantic retrieval methods, we introduce a structure-aware retrieval approach based on three key structural concepts: Conceptual Path, Syntactic Structure, and Causal Pattern. 
\begin{itemize}
    \item \textbf{Conceptual Path}: Captures the relationship between the source and target events, providing structural information through external knowledge.
    \item \textbf{Syntactic Structure}: Represents the syntactic properties of the context, commonly derived from dependency parsing and constituent parsing, offering insights into sentence structure.
    \item \textbf{Causal Pattern}: Proposed by \citet{cai-etal-2025-dr}, it consists of predefined simple causal graphs, providing pattern-matching-based structural information.
\end{itemize}

By incorporating the aforementioned structural concepts, we can more accurately retrieve examples that share the same reasoning structure and causal pattern. As illustrated in Figure~\ref{fig:example}, while the second and third instances exhibit little semantic similarity, they share similar Conceptual Paths, syntactic structures, and identical Causal Patterns, enabling the correct retrieval of the third instance.

To better measure these structural concepts, we propose \textbf{SERE}, a \textbf{S}tructural \textbf{E}xample \textbf{R}etrieval framework designed to \textbf{E}nhance LLMs’ performance on ECI tasks. Specifically, the \textbf{SERE} framework evaluates corpus samples using two key metrics: (i) Conceptual Path Metric: For each corpus sample and the target query, we extract the paths of their source event and target event from ConceptNet. The structural similarity is then measured using the \textit{edit distance} between these paths, producing a path-based relevance score. (ii) Syntactic Metric: We extract the syntactic trees of both the corpus sample and the target query, computing the \textit{tree edit distance} to quantify their structural alignment. The final score of each corpus sample is obtained by weighting and aggregating the scores from both metrics. Additionally, we introduce an LLM-based causal pattern extractor to identify and compare the causal patterns of both the corpus samples and the target query. Corpus samples that share the same causal pattern as the query and rank among the top-k highest scores are selected as the final examples. These retrieved examples are then incorporated into prompt design to guide the LLM in performing causal reasoning.

The main contributions of this paper are as follows: 
\begin{itemize}
   \item We propose \textbf{SERE}, a structural example retrieval framework that integrates three structural concepts—Conceptual Path, Syntactic Structure, and Causal Pattern—to enhance LLMs' ability to identify causal relationships in ECI tasks. To the best of our knowledge, we are the first to jointly integrate them for ECI task.
    \item We introduce two structural similarity metrics, \textbf{Conceptual Path Metric} and \textbf{Syntactic Metric}, along with a \textbf{Causal Pattern Filtering mechanism}, to effectively retrieve structurally aligned examples that guide LLM reasoning.
    \item We conduct extensive experiments on multiple ECI datasets, demonstrating the effectiveness, generalizability and robustness of \textbf{SERE}.
\end{itemize}

\begin{figure*}[!ht]
    \includegraphics[width=\linewidth]{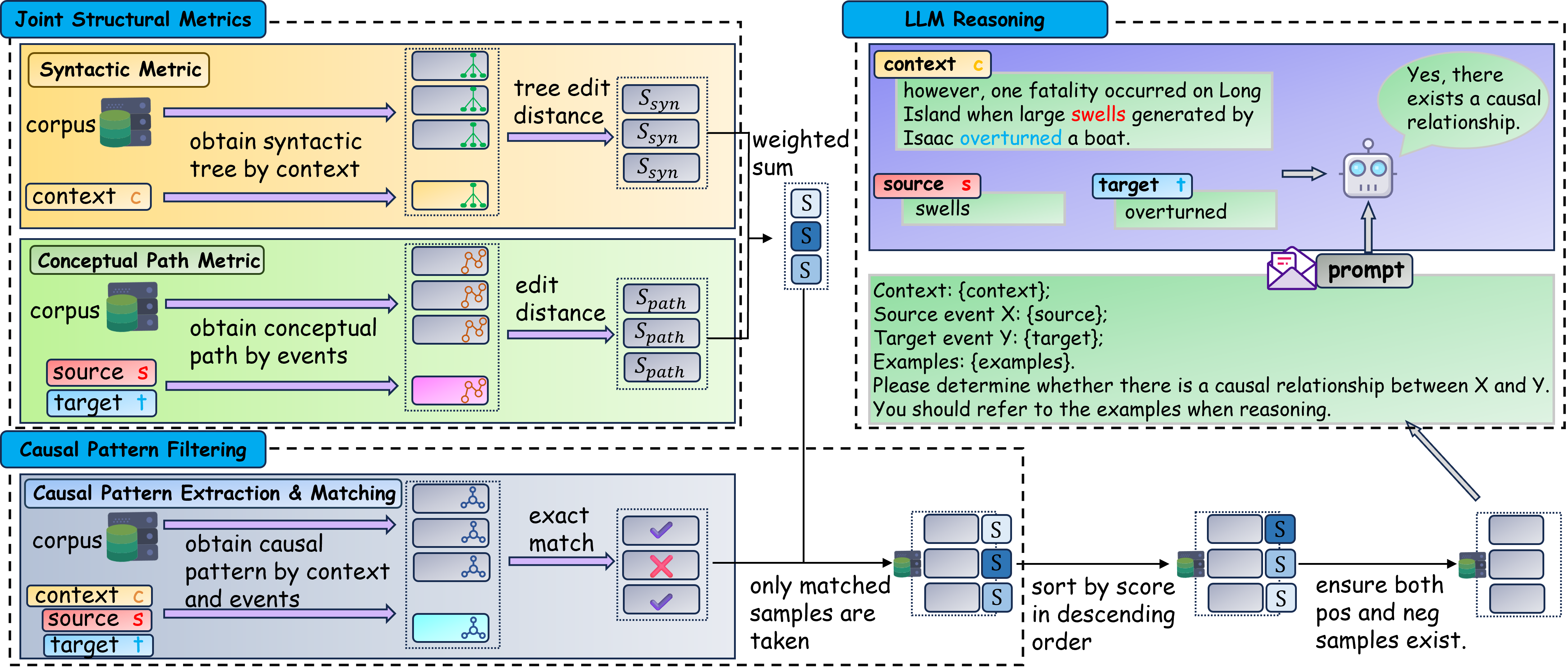}
    \caption{Overview of the SERE framework.}
    \label{fig:overview}
\end{figure*}

\section{Preliminary Introduction for Structures}
In this section, We introduce the three structural concepts used in this paper.

\textbf{Conceptual Path}: By connecting several concepts and the relationships between them, we can form a path between two events, known as a Conceptual Path. This path reveals the commonsense causal connections between events through a pre-constructed concept network, suitable for supplementing implicit connections between events missing in the text. It provides supplementary information to the model from the perspective of external knowledge priors. 

\textbf{Syntactic Structure}: This reveals the structural features of the context and provides causal clues at the syntactic level. It directly derives from the language structure of the text and can parse specific expressions, directionality, and complex syntactic patterns in the text, playing a role in causal reasoning from a syntactic prior perspective. 

\textbf{Causal Pattern}: It provides a structured representation of causal relationships by defining several causal structures that can be described by simple directed acyclic graphs (DAGs). This assists the model in causal reasoning from the perspective of structural pattern matching.

\section{Methodology}

\subsection{Overview}
The overview of the \textbf{SERE} framework is shown in Figure~\ref{fig:overview}. This framework consists of three components: \textit{Joint Structural Metrics}, \textit{Causal Pattern Filtering}, and \textit{LLM Reasoning}. In the \textit{Joint Structural Metrics} section, we compute the edit distance based score of the event pairs in ConceptNet for both the sample to be inferred and the corpus sample, as well as the tree edit distance based score of the syntactic trees of their contexts. After obtaining the two scores, we weight and sum them to generate the structure-based score for the corpus sample. In the \textit{Causal Pattern Filtering} section, we follow prior work and design an LLM Agent to generate the corresponding causal pattern for both the sample to be inferred and the corpus sample. We then select the top-k corpus samples that have the same causal pattern as the sample to be inferred and the highest structural scores as the final retrieved examples. In the \textit{LLM Reasoning} phase, we use the retrieved examples to design an appropriate prompt to guide the LLM in reasoning, ultimately obtaining the inference results.

\subsection{Notation Definition}
Let the sample to be inferred be $x = (c, s, t)$ and $y$, where $x$ represents the input, including $c$: the context string, $s$: the source event spans within $c$, and $t$: the target event spans within $c$; $y$: the ground truth causal label, which takes a value of \texttt{``yes''} or \texttt{``no''}. Additionally, we define the corpus as $\mathcal{Q} = \{(c_i, s_i, t_i, y_i) \mid i \in \mathbb{Z}^+\}$, where each sample also contains context, source event, target event, and the ground truth.

\subsection{Joint Structural Metrics}
In this section, we introduce two structure-based metrics to calculate the score of each sample in the corpus.

\subsubsection{Conceptual Path Metric}
This metric is based on ConceptNet \cite{speer2018conceptnet55openmultilingual}. ConceptNet is a freely available semantic knowledge graph, where the nodes represent various concepts from the real world and the edges represent the relationships between those concepts.

For a given sample $x = (c, s, t)$, we first match the concept nodes in ConceptNet corresponding to $s$ and $t$. To improve the matching rate, we use an encoder to encode both the events and all the nodes in ConceptNet. Then, we compute the cosine similarity between the event representation and the node representation, selecting the nodes with the highest similarity that exceeds a predefined threshold as the matching nodes for the events in ConceptNet, denoted as $node_s$ and $node_t$. Then, using the shortest path algorithm (denoted as $\text{SP}$), we find a path in ConceptNet between $s$ and $t$:
$$path_{s, t} = \text{SP}(node_s, node_t).$$

It is important to note that this path is not strictly unidirectional, but rather any path where there is an edge between any two nodes along the path. Using the same method, for each corpus sample $q_i \in \mathcal{Q}$, we obtain $path_{s_i, t_i}$. Subsequently, we calculate the edit distance ($\text{ED}$) based similarity between the two paths, obtaining the Conceptual Path based score:
$$S^{path}_{x, q_i} = 1 - \frac{\text{ED}(path_{s, t}, path_{s_i, t_i})}{\max(|path_{s, t}|, |path_{s_i, t_i}|)}, $$
where $|path_{s_i, t_i}|$ denotes the length of path between $s_i$ and $t_i$.

\subsubsection{Syntactic Metric}
This metric is based on the syntactic structure of the text. In this paper, we design our method using the dependency syntax tree as the syntactic feature of the text. For the given sample $x$'s text $c$, we first obtain its dependency syntax tree $tree_c$. Similarly, for each corpus sample $q_i \in \mathcal{Q}$, we obtain $tree_{c_i}$. In particular, for a text containing multiple sentences, we first obtain the syntax tree of each sentence separately, and then connect them to an artificial root node to construct the syntax tree of the entire text. After obtaining two trees, we calculate the tree edit distance ($\text{TED}$) based similarity between the two trees, obtaining the Syntax based score:

$$
S^{syn}_{x, q_i} = \mathrm{e}^{-0.05 \cdot \text{TED}(tree_c, tree_{c_i})}.
$$

After calculating $S^{path}_{x, q_i}$ and $S^{syn}_{x, q_i}$, we use their weighted sum as the joint structural score of the corpus sample $q_i$ for the sample to be inferred $x$: 
$$S_{x, q_i} = w_1 \cdot S^{path}_{x, q_i} + w_2 \cdot S^{syn}_{x, q_i},$$
where $w_1$ and $w_2$ are predefined weights. Using the weighted sum as the final score helps avoid missing samples that satisfy only specific structures.

\begin{algorithm}[h]
\small
\caption{Structure Based Example Retrieval}
\label{alg:example_retrievel}
\begin{algorithmic}[1]
\State \textbf{Input:} $x = (c, s, t)$; $\mathcal{Q} = \{(c_i, s_i, t_i, y_i)\}$; $k_{top}$

\Statex
\State $node_s \gets$ match ConceptNet node for $s$
\State $node_t \gets$ match ConceptNet node for $t$
\State $path_{s, t} \gets \text{SP}(node_s, node_t)$

\Statex
    
\State $tree_c \gets$ syntactic parse tree for $c$
\For{$q_i = (c_i, s_i, t_i, y_i) \in \mathcal{Q}$}
    \State $node_{s_i} \gets$ match ConceptNet node for $s_i$
    \State $node_{t_i} \gets$ match ConceptNet node for $t_i$
    \State $path_{s_i, t_i} \gets \text{SP}(node_{s_i}, node_{t_i})$
    \State $S^{path}_{x, q_i} \gets \text{norm\_sim}(\text{ED}(path_{s, t}, path_{s_i, t_i}))$

    \Statex
    
    \State $tree_{c_i} \gets$ syntactic parse tree for $c_i$
    \State $S^{syn}_{x, q_i} \gets \text{norm\_sim}(\text{TED}(tree_c, tree_{c_i}))$

    \Statex
    
    \State $S_{x, q_i} \gets w_1 \cdot S^{path}_{x, q_i} + w_2 \cdot S^{syn}_{x, q_i}$
\EndFor

\Statex
\State $cp \gets \text{PatternExtractor}(c, s, t)$
\State $\mathcal{E}' = \{\}$
\For{$q_i = (c_i, s_i, t_i, y_i) \in \mathcal{Q}$}
    \State $cp_i \gets \text{PatternExtractor}(c_i, s_i, t_i)$
    \If{$cp_i = cp$}
        \State $\mathcal{E}' \gets \mathcal{E}' \cup \{q_i\}$
    \EndIf
\EndFor

\Statex
\State $\mathcal{E}' \gets \text{sort } \mathcal{E}' \text{ by } S_{x, q_i} \text{ in descending order}$

\Statex
\State $k_{pos} \gets \lfloor k_{top} / 2 \rfloor$
\State $k_{neg} \gets k_{top} - k_{pos}$
\State $\mathcal{E}'_{pos} \gets \text{first } k_{pos} \text{ samples in } \{q_i \in \mathcal{E}' | y_i = \texttt{``yes''}\}$
\State $\mathcal{E}'_{neg} \gets \text{first } k_{neg} \text{ samples in } \{q_i \in \mathcal{E}' | y_i = \texttt{``no''}\}$
\State $\mathcal{E} \gets \mathcal{E}'_{pos} \cup \mathcal{E}'_{neg}$

\Statex
\State \textbf{Output:} $\mathcal{E}$
\end{algorithmic}
\end{algorithm}

\subsection{Causal Pattern Filtering}
In this section, we follow the prior work Dr.ECI \cite{cai-etal-2025-dr} to use the causal pattern as the structural feature of the instance and filtering samples based on this. Unlike Dr.ECI, we further specify the definition of causal patterns, allowing the LLM to more accurately distinguish between different patterns from a structural perspective, thereby achieving more precise pattern extraction. Moreover, in this step, we only require the LLM to predict the coarse-grained pattern according to predefined rules, which is easier for the model than directly performing end-to-end causal reasoning. The pattern descriptions are shown in Table~\ref{tb:causal_pattern}.

For a given sample $x = (c, s, t)$, we construct a prompt using its context and two events, allowing the LLM to generate the corresponding causal pattern, which is then extracted using regular expressions:
$$cp = \text{PatternExtractor}(c, s, t),$$
where $\text{PatternExtractor}$ is an prompted LLM. Similarly, for each positive sample in the corpus, we obtain its causal pattern:
$$cp_i = \text{PatternExtractor}(c_i, s_i, t_i).$$

For negative samples, their Causal Pattern is directly set to \texttt{``No''}.

Subsequently, we can begin filtering all samples in $\mathcal{Q}$ to obtain the examples needed for inference on $x$. First, we perform exact matching of the causal pattern to initially obtain 

$$\mathcal{E}' = \{q_i | q_i \in \mathcal{Q} \wedge cp_i = cp\}.$$

Following this, we sort $\mathcal{E}'$ in descending order based on $S_{x, q_i}$. We select $k_{top}$ examples with the highest scores from $\mathcal{E}'$. To reduce the ``causal hallucination'' phenomenon, we ensure that both positive and negative examples are selected. Finally, we get the resulting set of examples $\mathcal{E}$.

The algorithm for the entire example selection process is described in Algorithm~\ref{alg:example_retrievel}. It is worth noting that the overall retrieval pipeline is efficient and practical rather than complex, as long as the corpus and knowledge base are pre-constructed.

\begin{table}[!ht]
\begin{center}
\resizebox{0.5\textwidth}{!}{
\scalebox{0.2}{
\begin{tabular}{|c|c|}
    \hline %
        \textbf{Causal Pattern} 
        & \textbf{Causal Graph} \\
    \hline %
        \parbox[c]{2.5cm}{\centering Direct} 
        & \parbox[c]{2.5cm}{\vspace{0.5mm}\centering \begin{tikzpicture}[scale=0.5]
            \node (X) [circle, draw, scale=0.7] {\scalebox{0.7}{$X$}};
            \node (Y) [circle, draw, right of=X, node distance=1.5cm, scale=0.7] {\scalebox{0.7}{$Y$}};
            
            \draw[->] (X) -- (Y);
        \end{tikzpicture}\vspace{0.5mm}} \\
    \hline %
        \parbox[c]{2.5cm}{\centering Chain} 
        & \parbox[c]{2.5cm}{\vspace{0.5mm}\centering \begin{tikzpicture}[scale=0.5]
            \node (X) [circle, draw, scale=0.7] {\scalebox{0.7}{$X$}};
            \node (Y) [circle, draw, right of=X, node distance=1.5cm, scale=0.7] {\scalebox{0.7}{$Y$}}; %
            \node (Z) [circle, draw, below of=X, xshift=0.75cm, scale=0.7, node distance=0.65cm] {\scalebox{0.7}{$Z$}}; %
            
            \draw[->] (X) -- (Z);
            \draw[->] (Z) -- (Y);
            \draw[dashed,->] (X) to (Y); %
        \end{tikzpicture}\vspace{0.5mm}} \\
    \hline %
        \parbox[c]{2.5cm}{\centering Collider} 
        & \parbox[c]{2.5cm}{\vspace{0.5mm}\centering \begin{tikzpicture}[scale=0.5]
            \node (X) [circle, draw, scale=0.7] {\scalebox{0.7}{$X$}};
            \node (Y) [circle, draw, right of=X, node distance=1.5cm, scale=0.7] {\scalebox{0.7}{$Y$}}; %
            \node (Z) [circle, draw, below of=X, xshift=0.75cm, scale=0.7, node distance=0.65cm] {\scalebox{0.7}{$Z$}}; %
            
            \draw[->] (X) -- (Z);
            \draw[<-] (Z) -- (Y);
            \draw[dashed,->] (X) to (Y); %
        \end{tikzpicture}\vspace{0.5mm}} \\
    \hline %
        \parbox[c]{2.5cm}{\centering Fork} 
        & \parbox[c]{2.5cm}{\vspace{0.5mm}\centering \begin{tikzpicture}[scale=0.5]
            \node (X) [circle, draw, scale=0.7] {\scalebox{0.7}{$X$}};
            \node (Y) [circle, draw, right of=X, node distance=1.5cm, scale=0.7] {\scalebox{0.7}{$Y$}}; %
            \node (Z) [circle, draw, below of=X, xshift=0.75cm, scale=0.7, node distance=0.65cm] {\scalebox{0.7}{$Z$}}; %
            
            \draw[<-] (X) -- (Z);
            \draw[->] (Z) -- (Y);
            \draw[dashed,->] (X) to (Y); %
        \end{tikzpicture}\vspace{0.5mm}} \\
    \hline %
        \parbox[c]{2.5cm}{\centering Coreference}
        & \parbox[c]{2.5cm}{\vspace{0.5mm}\centering \begin{tikzpicture}[scale=0.5]
            \node (X) [circle, draw, scale=0.7] {\scalebox{0.7}{$X$}};
            \node (X') [circle, draw, below of=X, node distance=0.8cm, scale=0.7] {\scalebox{0.7}{$X'$}};
            \node (Y) [circle, draw, right of=X, xshift=0.5cm, scale=0.7] {\scalebox{0.7}{$Y$}};
            \node (Y') [circle, draw, below of=Y, node distance=0.8cm, scale=0.7] {\scalebox{0.7}{$Y'$}};
            
            \draw[->, dashed] (X) to (Y);
            \draw[-] (X) -- (X');
            \draw[-] (Y) -- (Y');
            \draw[->] (X) -- (Y');
            \draw[->] (X') -- (Y);
            \draw[->] (X') to (Y');
        \end{tikzpicture}\vspace{0.5mm}} \\
    \hline %
\end{tabular}
}
}
\end{center}
\caption{The Causal Patterns used in this paper. $X$ and $Y$ represent the source and target event, respectively; $Z$ represents mediators; $X'$ and $Y'$ denote the coreferential events of the corresponding events. Solid lines indicate explicitly stated relationships in the context, while dashed lines represent inferred implicit relationships. A more detailed description of the patterns can be found in the appendix~\ref{appendix:causal_pattern}.}\label{tb:causal_pattern}
\end{table}

\subsection{LLM Reasoning}
After obtaining $\mathcal{E}$, we can use it to guide the LLM for the final causal reasoning, which we treat as a few-shot ICL-based QA task. We design the prompt based on $x$ and $\mathcal{E}$, and use an LLM: $\text{Reasoner}$ for reasoning to obtain the final result: 
$$y' = \text{Reasoner}(x, \mathcal{E}),$$
where $y'$ is the final answer. Since this task is a binary classification task, $y'$ is restricted to be either \texttt{``Yes''} or \texttt{``No''}.

\section{Experiment}

\begin{table*}[ht]
\begin{center}
\resizebox{\textwidth}{!}{
\scalebox{0.2}{
\begin{tabular}{lccc|ccc|ccc}
    \toprule %
        \multirow{2}{*}{\textbf{Method}} & \multicolumn{3}{c|}{\textbf{ESC}} & \multicolumn{3}{c|}{\textbf{CTB}} & \multicolumn{3}{c}{\textbf{MAVEN-ERE}} \\
        & \textbf{P(\%)} & \textbf{R(\%)} & \textbf{F1(\%)} & \textbf{P(\%)} & \textbf{R(\%)} & \textbf{F1(\%)} & \textbf{P(\%)} & \textbf{R(\%)} & \textbf{F1(\%)} \\
    \midrule %
        \textbf{PaLM2:} & & & & & & & & & \\
        \hspace{1em} Dr.ECI\textsuperscript{\dag} & 29.0 & 75.4 & 41.9 & - & - & 13.0 & 22.0 & 80.0 & 34.5 \\
    \midrule %
        \textbf{GPT-4o-mini:} & & & & & & & & & \\
        \hspace{1em} Base & 28.4 & 82.9 & 42.3 & 5.4 & 80.5 & 10.1 & 23.1 & 91.5 & 36.9 \\
        \hspace{1em} CoT & 29.4 & 80.6 & 43.1 & 6.3 & 79.6 & 11.6 & 25.4 & 87.9 & 39.4 \\
        \hspace{1em} Dr.ECI & 31.0 & \textbf{90.2} & 46.1 & 8.2 & \textbf{91.2} & 15.1 & 26.0 & \textbf{93.0} & 40.7 \\ 
        \hspace{1em} SERE (ours) & \textbf{48.3} & 51.5 & \textbf{49.9} & \textbf{13.8} & 36.3 & \textbf{20.0} & \textbf{34.5} & 54.6 & \textbf{42.3} \\
        
    \midrule
        \textbf{Gemini-1.5-pro:} & & & & & & & & & \\
        \hspace{1em} Base & 24.0 & 80.8 & 37.0 & 4.7 & \textbf{95.6} & 9.0 & 21.2 & \textbf{95.1} & 34.6 \\
        \hspace{1em} CoT & 25.1 & \textbf{85.4} & 38.7 & 4.7 & 88.5 & 8.9 & 23.3 & 89.2 & 37.0 \\
        \hspace{1em} Dr.ECI & 27.6 & 82.0 & 41.3 & 7.2 & 88.5 & 13.3 & 23.4 & 91.7 & 37.3 \\ 
        \hspace{1em} SERE (ours) & \textbf{36.5} & 59.3 & \textbf{45.2} & \textbf{9.9} & 69.9 & \textbf{17.4} & \textbf{29.9} & 60.2 & \textbf{39.9} \\
    \bottomrule %
\end{tabular}
}
}

\end{center}
\caption{Main experiment results. The best scores are marked in bold. \dag: results from \cite{cai-etal-2025-dr}.}\label{tb:main_results}
\end{table*}

\subsection{Implementation Details}
For the Conceptual Path Metric, we construct a localized ConceptNet on Neo4j \cite{noauthororeditorneo4j}, encode nodes with Contriever-msmarco \cite{izacard2021contriever}, and query the shortest paths between events using Cypher. For the Syntactic Metric, we build dependency trees from the text using spaCy \cite{spacy2}.
The shortest path algorithm and distance-based similarity algorithm are provided in the appendix~\ref{appendix:shortest_path_alg} and~\ref{appendix:similarity_alg}.

In the main experiments, we set the confidence threshold for node matching to 0.6, the weights of the conceptual path and syntactic metrics ($w_1, w_2$) to 0.5 each, the number of selected examples ($k_{top}$) to 2, and the LLM temperature to 0 to ensure consistent responses.

We directly use the official APIs to access \texttt{gpt-4o-mini-2024-07-18} \cite{chatgpt} and \texttt{gemini-1.5-pro} \cite{geminiteam2024geminifamilyhighlycapable} in the main experiments.
The prompts used in this paper are provided in the appendi~\ref{appendix:prompts}.

\subsection{Datasets and Metrics}
We conduct experiments on three causality-annotated datasets: \textbf{EventStoryLine (ESC)} \cite{caselli-vossen-2017-event}, which provides event, temporal, and causality annotations (22 topics, 258 docs, 5,334 events, 1,770 causal pairs); \textbf{Causal-TimeBank (CTB)} \cite{mirza-etal-2014-annotating}, sourced from TempEval-3 \cite{uzzaman-etal-2013-semeval} (183 docs, 6,813 events, 318 causal pairs); and \textbf{MAVEN-ERE} \cite{wang-etal-2022-maven}, a large-scale Wikipedia-based dataset (90 topics, 4,480 docs, 103,193 events, 57,992 causal pairs). Data preprocessing follows \citet{gao2023chatgptgoodcausalreasoner}.
For evaluation, we adopt Precision (\textbf{P\%}), Recall (\textbf{R\%}), and F1 (\textbf{F1\%}) as metrics.

\subsection{Baselines}
In the main experiment, we primarily compare methods based on pre-trained LLMs:
\begin{itemize}
    \item \textbf{Base}: The LLM is prompted with a simple task description without additional guidance.
    \item \textbf{CoT} \cite{wei2023chainofthoughtpromptingelicitsreasoning}: The LLM is prompted to generate a step-by-step reasoning process before producing the final answer, enhancing its reasoning capability.
    \item \textbf{Dr.ECI} \cite{cai-etal-2025-dr}: The task is decomposed into multiple steps: the model first identifies mediators in the context that connect the target and source, and then follows predefined causal patterns to guide causal reasoning. For a fair comparison, we report both the original results obtained with PaLM2 \cite{anil2023palm2technicalreport} and our reproduced scores. However, since PaLM2 is not publicly accessible, we did not conduct experiments using PaLM2.
\end{itemize}

Since \textbf{SERE} relies on LLM APIs without fine-tuning, the main experiments only include non-fine-tuning baselines, while additional comparisons with fine-tuned models are provided in the analysis section.

To further demonstrate the generalizability of our method, we also conduct experiments under the fine-tuning setting, as detailed in the appendix~\ref{appendix:finetuned_llm}.

\subsection{Main Results}

\begin{table*}[h]
\begin{center}
\resizebox{\textwidth}{!}{
\scalebox{0.2}{
\begin{tabular}{lccc|ccc|ccc}
    \toprule %
        \multirow{2}{*}{\textbf{Method}} & \multicolumn{3}{c|}{\textbf{ESC-intra}} & \multicolumn{3}{c|}{\textbf{ESC-inter}} & \multicolumn{3}{c}{\textbf{CTB}} \\
        & \textbf{P(\%)} & \textbf{R(\%)} & \textbf{F1(\%)} & \textbf{P(\%)} & \textbf{R(\%)} & \textbf{F1(\%)} & \textbf{P(\%)} & \textbf{R(\%)} & \textbf{F1(\%)} \\
    \midrule %
        \hspace{1em} GenECI & 59.5 & 57.1 & 58.8 & - & - & - & 60.1 & 53.3 & 56.5 \\
        \hspace{1em} DPJL & 65.3 & 70.8 & 67.9 & - & - & - & 63.6 & 66.7 & 64.6 \\
        \hspace{1em} KEPT & 50.0 & 68.8 & 57.9 & - & - & - & 48.2 & 60.0 & 53.5 \\ 
        \hspace{1em} CPATT & 79.4 & 81.3 & \textbf{80.4} & 74.9 & 60.1 & 66.7 & 77.5 & 73.2 & 75.2 \\ 
        \hspace{1em} SERE (ours) & 75.5 & 81.6 & 78.4 & 66.7 & 87.2 & \textbf{75.6} & 84.3 & 89.4 & \textbf{86.8} \\
    \bottomrule %
\end{tabular}
}
}

\end{center}
\caption{CPATT settings results. The results of GenECI \cite{man-etal-2022-event}, DPJL \cite{shen-etal-2022-event}, KEPT \cite{LIU2023110064} and CPATT \cite{DBLP:journals/kbs/ZhangKZLMLW23} are all taken from the paper of CPATT. ESC-intra and ESC-inter refer to subsets of the ESC dataset where the two events appear in the same sentence or in different sentences, respectively. CTB does not distinguish between intra and inter cases. The best scores are marked in bold.}\label{tb:cpatt_results}
\end{table*}

We report the Precision, Recall, and F1 scores of different methods on the ESC, CTB, and MAVEN-ERE datasets, as detailed in Table~\ref{tb:main_results}. 

As shown in the table, our method achieves the highest F1 scores across all three datasets, demonstrating its generalization capability. The Base method for both models performs the worst on the CTB dataset, whereas our method achieves nearly double the score improvement and an approximately 20\% increase on the ESC dataset. Compared to the Base method, the CoT method shows slight improvements in most cases, but the gains are limited. CoT may help reduce the model’s reliance on surface-level causal keywords, thereby improving precision, but step-by-step reasoning alone is insufficient to address causal bias. The low Precision and high Recall of the Base and CoT methods indicate the presence of ``Causal Hallucination'', where the model without specific optimization tends to misclassifies non-causal relationships as causal. In contrast, \textbf{SERE} effectively suppresses such misclassifications, reducing the hallucination.

Dr.ECI generally exhibits higher Recall but limited improvements in Precision. In contrast, \textbf{SERE} enhances final performance by improving precision, indicating that our method is more conservative. Due to the use of three retrieval methods, \textbf{SERE} applies stricter criteria when selecting examples, which may lead to missing some causal relationships. This could explain why \textbf{SERE} achieves higher Precision but lower Recall. Considering that current LLM-based approaches struggle to balance precision and recall, our method achieves a trade-off between the two. While this means that some true causal relations may be missed, it significantly reduces the risk of the model generating spurious causal links, making our approach more suitable for risk-sensitive scenarios. Overall, \textbf{SERE} achieves stronger performance across different models and datasets.

\section{Analysis Experiments}

\subsection{CPATT Settings}

To demonstrate the generality of \textbf{SERE}, we further evaluate its performance in the fine-tuning scenario, following the experimental settings of CPATT \cite{DBLP:journals/kbs/ZhangKZLMLW23}, a prior state-of-the-art method. CPATT fine-tunes an LLM with a different preprocessing strategy from \citet{gao2023chatgptgoodcausalreasoner}: for long contexts, only sentences containing the target events are retained, while negative samples are constructed by randomly pairing non-causal events with their corresponding sentences. We directly perform inference on the test split.

As shown in Table~\ref{tb:cpatt_results}, \textbf{SERE} performs consistently well. On ESC-intra, it achieves an F1 of 78.4\%, ranking second only to CPATT, while on ESC-inter it attains the best F1 of 75.6\%. Although fine-tuned CPATT performs better in intra-sentence settings, its precision and recall drops in cross-sentence contexts. In contrast, the non-fine-tuned SERE remains stable across settings, showing stronger robustness and generalization. On CTB, \textbf{SERE} also achieves marginally the highest precision, recall, and F1, further confirming its cross-domain robustness.

Note that CPATT’s preprocessing is primarily designed for training, where random negative sampling helps reduce causal hallucination; therefore, these results are not included in the main experiments. Nevertheless, they clearly demonstrate the strong generalization ability of \textbf{SERE}.

\subsection{Ablation Study}

\begin{table}[h]
\resizebox{0.5\textwidth}{!}{
\scalebox{0.8}{
\begin{tabular}{lccc}
    \toprule %
        \textbf{Component} & \textbf{ESC} & \textbf{CTB} & \textbf{MAVEN-ERE} \\
    \midrule %
        \textbf{SERE} & 49.9 & 20.0 & 42.3 \\
        \textbf{w/ Conceptual Path} & 46.7 & 18.9 & 39.2 \\
        \textbf{w/ Syntactic} & 44.5 & 18.9 & 38.7 \\
        \textbf{w/ Causal Pattern} & 47.1 & 17.3 & 38.3 \\
    \bottomrule %
\end{tabular}
}
}
\caption{Performance when retrieval is guided by a single structural signal. For ``w/ Causal Pattern'', we randomly select k samples with the same pattern. The evaluation metric is F1(\%).}\label{tb:analysis_with_structure}
\end{table}

\begin{table}[h]
\resizebox{0.5\textwidth}{!}{
\scalebox{0.8}{
\begin{tabular}{lccc}
    \toprule %
        \textbf{Component} & \textbf{ESC} & \textbf{CTB} & \textbf{MAVEN-ERE} \\
    \midrule %
        \textbf{SERE} & 49.9 & 20.0 & 42.3 \\
        \textbf{w/o Conceptual Path} & 46.6 & 18.8 & 40.8 \\
        \textbf{w/o Syntactic} & 48.1 & 18.9 & 40.5 \\
        \textbf{w/o Causal Pattern} & 47.5 & 18.2 & 39.4 \\
    \bottomrule %
\end{tabular}
}
}
\caption{Performance when one structural signal is removed from \textbf{SERE}. For ``w/o Causal Pattern'', the Causal Pattern Filtering step is skipped, and retrieval relies solely on the Joint Structural Metrics. The evaluation metric is F1 (\%).}\label{tb:analysis_without_structure}
\end{table}

We conduct ablation studies to examine the role of three structural signals in retrieval. As shown in Table~\ref{tb:analysis_with_structure}, using any single structural signal alone leads to a noticeable performance drop compared to the full \textbf{SERE} model, although it still outperforms the Base method. This suggests that each signal captures useful but incomplete structural information.

Further analysis in Table~\ref{tb:analysis_without_structure} shows that removing any one component from \textbf{SERE} also results in consistent degradation across all datasets. This indicates that the three signals provide complementary, non-redundant contributions to the retrieval process. While combining any two signals already yields clear improvements over the Base method, it remains insufficient to match the full model.

Overall, the best performance is achieved when all three structural signals are jointly incorporated. This demonstrates that effective retrieval benefits from simultaneously enforcing structural type consistency (via Causal Pattern) and structural similarity (via Conceptual path and syntactic metrics), leading to more accurate and robust example selection.

\subsection{Alternative Retrieval Methods Analysis}

\begin{table}[h]
\resizebox{0.5\textwidth}{!}{
\scalebox{0.8}{
\begin{tabular}{lccc}
    \toprule %
        \textbf{Retrieval Method} & \textbf{ESC} & \textbf{CTB} & \textbf{MAVEN-ERE} \\
    \midrule %
        \textbf{Base} & 42.3 & 10.1 & 36.9 \\
        \textbf{Random} & 46.6 & 13.9 & 39.1 \\
        \textbf{Contriever-msmarco} & 46.2 & 18.8 & 37.9 \\
        \textbf{BM25} & 46.5 & 17.1 & 33.4 \\
    \bottomrule %
\end{tabular}
}
}
\caption{Experiments of three retrieval methods. The evaluation metric is F1 (\%).
}\label{tb:analysis_retriever}
\end{table}

We evaluate three retrieval strategies. \textbf{Random} samples k instances uniformly at random. \textbf{Contriever-msmarco} encodes text into dense vectors and retrieves examples based on cosine similarity. \textbf{BM25} ranks candidates using term frequency–inverse document frequency (TF–IDF). In both cases, the top-k samples are selected according to their relevance scores. The results are shown in Table~\ref{tb:analysis_retriever}. Compared with \textbf{SERE}, all three reported methods (Random, Contriever-msmarco, and BM25) lead to performance drops. Although adding examples generally helps—yielding higher scores than the Base method on ESC and CTB—BM25 performs worse than Base on MAVEN-ERE, indicating that traditional scoring-based retrieval may be unsuitable for ECI.

By jointly examining Tables~\ref{tb:analysis_with_structure} and~\ref{tb:analysis_without_structure}, we observe that using a single structural signal or a pair of signals generally performs better than traditional retrieval, underscoring the need to integrate structural cues.

\subsection{Effect of varying the number of retrieved examples}

\begin{table}[!h]
\resizebox{0.5\textwidth}{!}{
\scalebox{0.8}{
\begin{tabular}{lccc}
    \toprule %
        \textbf{top-k} & \textbf{ESC} & \textbf{CTB} & \textbf{MAVEN-ERE} \\
    \midrule %
        \textbf{top-2} & 49.9 & 20.0 & 42.3 \\
        \textbf{top-4} & 49.0 & 22.7 & 39.8 \\
        \textbf{top-6} & 47.1 & 19.7 & 38.9 \\
    \bottomrule %
\end{tabular}
}
}
\caption{The results of varying numbers of retrieved examples. The evaluation metric is F1 (\%).}\label{tb:top-k}
\end{table}

We study the impact of varying the number of retrieved examples on the performance of \textbf{SERE}. In the main experiment, we set $k_{top} = 2$. As shown in Table~\ref{tb:top-k}, increasing $k_{top}$ to 4 yields a slight improvement on CTB, but leads to minor performance drops on ESC and MAVEN-ERE. When $k_{top}$ is further increased to 6, performance consistently degrades across all three datasets.

We attribute this trend to the limited context capacity of LLMs. Incorporating more demonstrations increases the prompt length and may introduce structurally similar yet less relevant examples, which can dilute informative signals. Notably, across all examined values, \textbf{SERE} consistently outperforms Base, CoT, and Dr.ECI, indicating that the method remains robust within a reasonable range of $k_{top}$.

\subsection{Causal Pattern Extraction Analysis}

\begin{figure}[!h]
    \centering
    \includegraphics[width=0.9\linewidth]{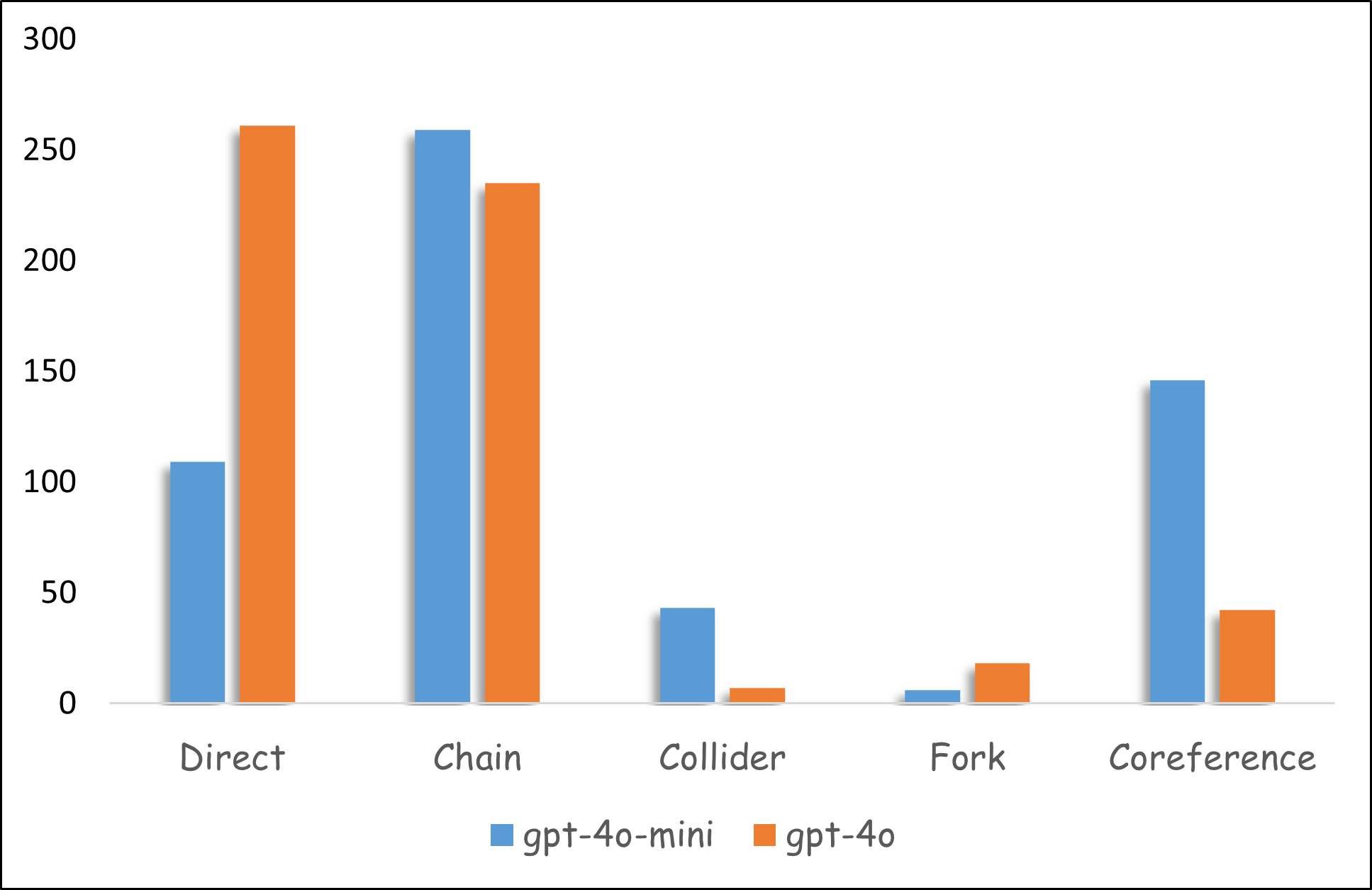}
    \caption{The Causal Pattern extraction results for the positive samples in the ESC dataset by two models. Due to cost constraints, we conduct experiments only on a subset of ESC.}
    \label{fig:analysis_pattern}
\end{figure} 

We analyze the Causal Pattern extraction results of GPT-4o-mini and GPT-4o on the positive sample set of the ESC dataset, as shown in Figure~\ref{fig:analysis_pattern}.

Both models identify a large number of Direct and Chain patterns, while Collider and Fork patterns are rarely recognized. This suggests that the ESC dataset is likely dominated by the first two patterns, which aligns with intuition: most texts, especially the short texts in the dataset, contain relatively simple causal relationships or are explicitly marked by causal trigger words. The low occurrence of the latter two patterns may also indicate that the models struggle to recognize more complex causal structures.

GPT-4o perceives Direct and Chain patterns in similar proportions, whereas GPT-4o-mini identifies significantly fewer Direct patterns compared to Chain. This suggests that the mini one may have a stronger preference for parsing multi-hop reasoning while being less inclined to recognize direct causal relationships. Additionally, some causal relationships that the mini model considers to require intermediate steps are classified as Direct by GPT-4o. This may be due to GPT-4o’s stronger reasoning ability, allowing it to infer causal relationships directly based on its internal knowledge.

\section{Related Work}
\subsection{Structure-based Methods in ECI}

Traditional structure-based methods mainly focus on lexical and syntactic patterns \cite{riaz-girju-2013-toward, hashimoto-etal-2014-toward}, aiming to capture surface-level cues that signal causal relations. Rule-based approaches typically rely on handcrafted linguistic patterns and predefined causal connectors (\eg, \texttt{``causes''}, \texttt{``leads to''}), which offer interpretability but require substantial manual effort and domain expertise. 

To alleviate these issues, recent studies have increasingly turned to pre-trained models, such as SemSIn \cite{hu2023semanticstructureenhancedevent} for implicit causal detection and Dr.ECI \cite{cai-etal-2025-dr} for predefined causal patterns. In addition, some works incorporate external knowledge bases such as ConceptNet \cite{cao-etal-2021-knowledge, huang-etal-2024-distill, LIU2023110064, su-etal-2025-enhancing} to facilitate the discovery of latent relations between events. However, most existing research relies on training models, which increases implementation complexity and limits the flexibility of these approaches.

\subsection{Generative LLM in ECI}

The rise of generative LLMs has reshaped the ECI research paradigm. Many methods that directly invoke LLMs have achieved strong performance in the ECI domain \cite{sun2024eventcausalitykeycomputational, wang-etal-2024-event-causality, guan-etal-2025-mmd, zeng2025zeroshoteventcausalityidentification}, demonstrating the potential of large language models to capture complex causal relations without extensive task-specific training.

However, despite their causal reasoning ability, the “causal hallucination” phenomenon highlights a clear gap compared with fine-tuned models \cite{gao2023chatgptgoodcausalreasoner}, particularly in tasks requiring precise and reliable inference.

Many prompt-based fine-tuning methods for LLMs have also demonstrated significant effectiveness on the ECI task \cite{peng2023doesincontextlearningfall, hu-etal-2025-large, xiang-etal-2025-evaluating}. By incorporating task-specific prompts or instructions during training, these approaches can better align the model with causal reasoning objectives. Nevertheless, they still rely heavily on the availability of high-quality annotated data, and their performance can degrade when applied to low-resource or cross-domain settings.

\subsection{In-context Learning in LLM}

The in-context learning (ICL) ability of LLMs allows them to perform tasks with only a few examples, as widely demonstrated in prior work \cite{brown2020languagemodelsfewshotlearners, touvron2023llamaopenefficientfoundation, bertsch2025incontextlearninglongcontextmodels, li2024longcontextllmsstrugglelong}. However, ICL performance is highly sensitive to factors such as the number, order, distribution, and quality of examples \cite{agarwal2024manyshotincontextlearning, kumar2021reorderingexampleshelpsprimingbased, min2022rethinkingroledemonstrationsmakes, zhao2021calibrateuseimprovingfewshot, arora2022askanythingsimplestrategy, voronov2024mindformatconsistentevaluation}, making the selection and processing of samples crucial \cite{peng2024revisitingdemonstrationselectionstrategies}. Traditional approaches often rely only on general statistical patterns or sentence semantics and do not exploit task-specific causal priors, which limits their suitability for ECI. Therefore, we propose a structure-based method.

\section{Conclusion}
In this paper, we propose \textbf{SERE}, a structural example retrieval framework designed to enhance LLM performance in ECI tasks. \textbf{SERE} introduces three structural factors: conceptual path, syntactic structure, and causal patterns. The core idea of \textbf{SERE} is to integrate these structural concepts and design a retrieval method that incorporates two structural similarity metrics—Conceptual Path Metric and Syntactic Metric—along with a Causal Pattern Filtering mechanism into few-shot learning, thereby improving LLMs' causal reasoning ability in ECI.  
Experimental results demonstrate the effectiveness of \textbf{SERE} across multiple ECI datasets. We believe that \textbf{SERE} and the underlying structural concepts provide valuable insights for future research in ECI, particularly in refining example selection strategies and improving LLM adaptability in causal reasoning tasks.

\section*{Limitations}
This work has two main limitations: (1) Due to the high cost of API calls and deploying, we did not evaluate the performance of other mainstream large models or more open-source models. (2) For the three types of structural information mentioned in this paper, we only explore their application in example retrieval, and further development remains for future research.

\section*{Acknowledgments}
This research was supported in part by National Science and Technology Major Project (2021ZD0111502), Natural Science Foundation of China (U24A20233, 62406078, 62476163), the Guangdong Basic and Applied Basic Research Foundation (2023B1515120020), CCF-DiDi GAIA Collaborative Research Funds (CCF-DiDi GAIA 202521), and Guangdong Laboratory of Artificial Intelligence and Digital Economy (SZ)(GML-KF-24-23).

\bibliography{custom}

\appendix

\section{Broader Societal Implications and Potential Risks}

Potential societal risks primarily relate to biases in external knowledge sources and the consequences of incorrect causal inference in high-stakes contexts. External knowledge bases such as ConceptNet may encode cultural, gender, or socio-demographic biases and exhibit uneven domain coverage. If not carefully managed, these biases could influence retrieval behavior or subtly distort causal interpretation. In addition, misidentifying causal relations in domains such as healthcare, law, or public policy may lead to misleading conclusions or inappropriate downstream decisions.

\textbf{SERE} mitigates these risks by not treating any single source—whether ConceptNet relations or LLM outputs—as authoritative causal evidence. Instead, these components serve as complementary structural cues within a retrieval-based framework alongside syntactic structure and causal pattern abstractions. This multi-signal design helps prevent any single biased or spurious source from dominating the final decision, and grounding predictions in multiple retrieved examples further reduces the impact of isolated errors.

Even so, responsible use of event causal inference systems requires continued attention to fairness, robustness, and domain safety. Future work may incorporate bias detection, domain-aware filtering of external knowledge, and human oversight, particularly in safety-critical applications.

\section{Performance of Finetuned LLM}
\label{appendix:finetuned_llm}

\begin{table*}[t]
\begin{center}
\resizebox{\textwidth}{!}{
\scalebox{0.2}{
\begin{tabular}{lccc|ccc|ccc|ccc ccc}
	\toprule %
		\multirow{2}{*}{\textbf{Method}} & \multicolumn{3}{c|}{\textbf{ESC-intra}} & \multicolumn{3}{c|}{\textbf{ESC-inter}} & \multicolumn{3}{c|}{\textbf{ESC}} & \multicolumn{3}{c}{\textbf{CTB}} \\
		& \textbf{P(\%)} & \textbf{R(\%)} & \textbf{F1(\%)} & \textbf{P(\%)} & \textbf{R(\%)} & \textbf{F1(\%)} & \textbf{P(\%)} & \textbf{R(\%)} & \textbf{F1(\%)} & \textbf{P(\%)} & \textbf{R(\%)} & \textbf{F1(\%)} \\
	\midrule %
		\textbf{CPATT} & 79.4 & 81.3 & 80.4 & 74.9 & 60.1 & 66.7 & 76.5 & 66.2 & 71.0 & 77.5 & 73.2 & 75.2 \\
		\textbf{Qwen2.5-3B-Inst:} & & & & & & & & & & & & \\
        
		\hspace{1em} \textbf{Base} & 76.1 & 84.3 & 79.9 & 77.3 & 66.7 & 71.6 & 78.5 & 69.4 & 73.7 & 95.0 & 86.4 & 90.5 \\
		\hspace{1em} \textbf{SERE} & \textbf{77.4} & \textbf{90.1} & \textbf{83.3} & \textbf{78.8} & \textbf{72.8} & \textbf{75.7} & \textbf{79.3} & \textbf{73.7} & \textbf{76.4} & \textbf{91.3} & \textbf{95.5} & \textbf{93.3} \\
	\bottomrule %
\end{tabular}
}
}
\end{center}
\caption{Comparison between finetuned \textbf{SERE} and CPATT. Base indicates that the fine-tuning instructions do not include retrieved examples; \textbf{SERE} indicates that the instructions include examples retrieved by the \textbf{SERE} method. The results of CPATT are taken from the original paper \cite{DBLP:journals/kbs/ZhangKZLMLW23}.}\label{tb:finetuned_llm}
\end{table*}

To further demonstrate the generalizability of the \textbf{SERE} method, we fine-tune Qwen2.5-3B-Inst \cite{qwen2025qwen25technicalreport} using the supervised instruction-based LoRA \cite{hu2021loralowrankadaptationlarge} approach with the LlamaFactory library \cite{zheng2024llamafactory}. Each training instance has the same input as in the API-based setup. We then compare it with CPATT, which is also fine-tuned (using BART-base \cite{DBLP:journals/corr/abs-1910-13461} with 139M parameters). The results are shown in Table~\ref{tb:finetuned_llm}.

As the results show, \textbf{SERE} that uses fine-tuned Qwen outperforms both its base counterpart and the CPATT method in terms of F1 score, achieving a new state-of-the-art level. In essence, fine-tuning improves both precision and recall, thereby alleviating the issue of ``causal hallucination'' while ensuring maximum coverage of positive instances.

It should be noted that our \textbf{SERE} method is not originally designed for the fine-tuning setting. Therefore, we do not consider this experiment part of the main experiments. Nevertheless, based on these scores, we believe this experiment demonstrates the generalizability and robustness of the \textbf{SERE} method in the fine-tuning setting.

\section{Effects of Individual Structural Components}

\begin{table*}[t]
\begin{center}
\resizebox{\textwidth}{!}{
\scalebox{0.2}{
\begin{tabular}{lccc|ccc|ccc}
    \toprule %
        \multirow{2}{*}{\textbf{Method}} & \multicolumn{3}{c|}{\textbf{ESC}} & \multicolumn{3}{c|}{\textbf{CTB}} & \multicolumn{3}{c}{\textbf{MAVEN-ERE}} \\
        & \textbf{P(\%)} & \textbf{R(\%)} & \textbf{F1(\%)} & \textbf{P(\%)} & \textbf{R(\%)} & \textbf{F1(\%)} & \textbf{P(\%)} & \textbf{R(\%)} & \textbf{F1(\%)} \\
    \midrule %
        \textbf{Base} & 28.4 & 82.9 & 42.3 & 5.4 & 80.5 & 10.1 & 23.1 & 91.5 & 36.9 \\
        \textbf{w/ Conceptual Path} & 28.8 & 93.3 & 44.0 & 5.7 & 93.8 & 10.7 & 24.0 & 92.6 & 38.1 \\
        \textbf{w/ Syntactic Tree} & 29.3 & 91.1 & 44.3 & 6.0 & 93.8 & 11.3 & 25.5 & 91.1 & 39.9 \\
        \textbf{w/ Causal Pattern} & 31.6 & 75.5 & 44.5 & 6.4 & 72.6 & 11.8 & 26.4 & 77.8 & 39.5 \\
    \bottomrule %
\end{tabular}
}
}
\end{center}
\caption{Effects of individual structural components (Conceptual Path, Syntactic Tree and Causal Pattern) when added to the Base prompt.}\label{tb:individual_structure_in_base}
\end{table*}

To further support our design and clarify the effect of each structure, we conduct additional ablation studies. We directly incorporate Conceptual Path, Syntactic Tree, and Causal Pattern into the Base prompt, without introducing examples. As shown in table~\ref{tb:individual_structure_in_base}, all three structures independently improve the model’s performance compared to the Base prompt. Conceptual Path significantly boosts Recall, especially on ESC and CTB. This aligns with intuition—LLMs often tend to predict "causal" by default, and the path connecting events makes the model more confident, albeit at the risk of false positives. Syntactic Tree also improves both Precision and Recall slightly, possibly because it helps the model better understand grammatical dependencies relevant to causality. Causal Pattern enhances Precision but reduces Recall. We attribute this to the stricter constraints imposed by predefined patterns, which encourage the model to avoid overpredicting causality—thereby mitigating hallucination at the cost of coverage.

These results further validate that each structure contributes differently and meaningfully to the model’s reasoning behavior. While our current work applies these signals in a retrieval scenario, we believe their independent effects suggest deeper potential.

\section{Applying Dr.ECI's Causal Pattern in \textbf{SERE}}

\begin{table}[!h]
\resizebox{0.5\textwidth}{!}{
\scalebox{0.8}{
\begin{tabular}{lccc}
    \toprule %
        \textbf{Method} & \textbf{ESC} & \textbf{CTB} & \textbf{MAVEN-ERE} \\
    \midrule %
        \textbf{SERE} & 49.9 & 20.0 & 42.3 \\
        \textbf{w/ Dr.ECI} & 47.0 & 16.1 & 41.4 \\
    \bottomrule %
\end{tabular}
}
}
\caption{The results of applying Dr.ECI's Causal Patterns to \textbf{SERE}. The evaluation metric is F1 (\%).}\label{tb:dreci_pattern}
\end{table}

We conduct experiments using the original Causal Pattern defined in the Dr.ECI paper \cite{cai-etal-2025-dr} on \textbf{SERE}, and the results are shown in Table~\ref{tb:dreci_pattern}. As can be seen, using the original causal pattern leads to a performance drop for \textbf{SERE} across all three datasets. This may be due to the inclusion of some rules unrelated to structure in the extraction process of the causal pattern, such as determining direct/indirect pattern based on the presence of the trigger words, and allowing the LLM to determine the pattern based on common sense. This causes the model to inaccurately grasp the intrinsic structure of the sample, leading to incorrect pattern extraction and ultimately misleading the final causal reasoning.

\section{Inference Cost Analysis}

\begin{table}[!h]
\resizebox{0.5\textwidth}{!}{
\scalebox{0.8}{
\begin{tabular}{lccc}
    \toprule %
        \textbf{Method} & \textbf{Time (s)} & \textbf{Input Tokens} & \textbf{Output Tokens}\\
    \midrule %
        \textbf{CoT} & 4.02 & 137 & 285 \\
        \textbf{Dr.ECI} & 14.26 & 2017 & 872 \\
        \textbf{SERE} & 21.85 & 1843 & 621 \\
    \bottomrule %
\end{tabular}
}
}
\caption{The average time and the token usage of different methods during inference.}\label{tb:inference_cost}
\end{table}

To provide a clearer picture of \textbf{SERE}’s efficiency, we conduct an additional runtime and token usage analysis using 300 random samples from the ESC dataset. We measure inference latency and token cost using GPT-4o-mini for the three methods: CoT, Dr.ECI, and \textbf{SERE}. The results are shown in table \ref{tb:inference_cost}.

\textbf{SERE} is slower overall primarily because it performs structural example retrieval over a corpus of 3,000 instances. However, only about 8 seconds of \textbf{SERE}’s runtime come from LLM inference. The remaining time is dominated by structure-matching computations—especially tree-edit-distance evaluation for syntactic structures—which are performed on CPU.

Although structural retrieval introduces additional overhead, \textbf{SERE} achieves substantially better task performance. Given that the retrieval step can naturally benefit from standard acceleration strategies such as parallel computation, cached structural features in future deployments, we believe this cost–effectiveness trade-off is acceptable for many settings.

In terms of token consumption, \textbf{SERE} consumes fewer tokens than Dr.ECI, since it uses fewer LLM calls and retrieves a controlled number of demonstrations. This makes \textbf{SERE} comparatively more practical in environments where token cost is a bottleneck.

\section{Computational Cost and Baseline Comparison}

\begin{table}[!h]
\resizebox{0.5\textwidth}{!}{
\scalebox{0.8}{
\begin{tabular}{lccc}
    \toprule %
        \textbf{Method} & \textbf{ESC} & \textbf{CTB} & \textbf{MAVEN-ERE} \\
    \midrule %
        \textbf{Stage 1} & 43.7 & 11.3 & 39.3 \\
        \textbf{Stage 2 (w/ examples)} & 12.5 & 1.6 & 11.7 \\
        \textbf{Stage 2 (w/o examples)} & 31.6 & 7.3 & 29.9 \\
        \textbf{SERE} & \textbf{49.9} & \textbf{20.0} & \textbf{42.3} \\
    \bottomrule %
\end{tabular}
}
}
\caption{Comparison of \textbf{SERE} with two-stage LLM refinement baselines.}\label{tb:llm_calls_baseline}
\end{table}

We quantify the number of LLM calls and assess whether comparable performance can be achieved using simpler inference pipelines. In \textbf{SERE}, LLM usage mainly occurs in two stages: (1) a one-time corpus preprocessing step, where each instance is processed once to extract Causal Patterns, and (2) the inference stage, where each test instance requires two calls—one for extracting its Causal Pattern and one for generating the final prediction conditioned on retrieved examples.

To investigate whether similar performance can be obtained with fewer calls, we construct a two-stage refinement baseline. In Stage 1, the LLM predicts causal relations using a CoT prompt. In Stage 2, it is queried again to reassess the initial reasoning and output the final label, optionally with randomly selected examples. This setup limits inference to two LLM calls per instance, acknowledging that corpus-level pattern extraction in \textbf{SERE} is performed once offline and does not add to per-sample cost.

The results in Table \ref{tb:llm_calls_baseline} show that invoking the LLM twice without carefully curated evidence leads to a substantial performance drop. The degradation becomes even more pronounced when randomly selected examples are introduced, indicating that the model cannot reliably self-correct using unrelated or unstructured contextual inputs. These results demonstrate that \textbf{SERE}’s improvements do not simply stem from increasing the number of LLM calls or exposing the model to arbitrary examples.

\section{Case Study}

\begin{figure*}[t]
\centering
\begin{case-box}[label={case:cot_and_sere}]{Handling Hallucination}

  \begin{tcolorbox}[colback=gray!3, title=Input, fonttitle=\bfseries]
    Williams was \textbf{[event2: booked into]} county jail and later released on \textbf{[event1: bail]}.
  \end{tcolorbox}

  \begin{tcolorbox}[colback=gray!3, title=CoT, fonttitle=\bfseries]
    To determine if there is a causal relationship between Event X (bail) and Event Y (booked into), I will analyze the text provided.
    \vspace{1em}

    \hspace{1em} \textbf{1.} The text states that "Williams was booked into county jail and later released on bail."
    
    \hspace{1em} \textbf{2.} This indicates that the booking into jail (Event Y) occurred first, followed by the release on bail (Event X).
    
    \hspace{1em} \textbf{3.} The sequence of events suggests that being booked into jail is a prerequisite for being released on bail. Therefore, the booking into jail leads to the possibility of being released on bail.
    
    \hspace{1em} \textbf{4.} This implies a causal relationship where Event Y (booked into) leads to Event X (bail).
    \vspace{1em}
    
    Based on this reasoning, I conclude that there is a causal relationship where Y -> X.
    \vspace{1em}
    
    Final answer in JSON format:
    
    \textbf{\{"Answer": "Yes"\}}
  \end{tcolorbox}

  \begin{tcolorbox}[colback=gray!3, title=SERE, fonttitle=\bfseries]
    To determine whether there is a causal relationship between Event X (bail) and Event Y (booked into), I will follow a step-by-step reasoning process:
    \vspace{1em}

    \textbf{1. **Identify the Events**:}
       \hspace{1em} - Event X: "bail" refers to the process of being released from custody after paying a certain amount of money.
       
       \hspace{1em} - Event Y: "booked into" refers to the act of being formally admitted into a jail or prison.
       \vspace{1em}
    
    \textbf{2. **Analyze the Context**:}
       \hspace{1em} - The text states that Williams was "booked into county jail" and later "released on bail." This indicates a sequence of events where the booking into jail occurs first, followed by the release on bail.
       \vspace{1em}
    
    \textbf{3. **Determine the Nature of the Relationship**:}
       \hspace{1em} - The act of being "booked into" jail is a prerequisite for the possibility of being released on bail. In other words, one cannot be released on bail without first being booked into jail.
       
       \hspace{1em} - However, the act of "bail" itself does not cause the booking into jail; rather, it is a consequence of being booked.
       \vspace{1em}
    
    \textbf{4. **Causal Direction**:}
       \hspace{1em} - The causal direction is important here. The booking into jail (Event Y) is what allows for the possibility of bail (Event X). Therefore, while bail is a result of the booking, it does not cause the booking.
       \vspace{1em}
    
    \textbf{5. **Conclusion**:}
       \hspace{1em} - Since Event Y (booked into) must occur before Event X (bail) can happen, and Event X does not influence or cause Event Y, there is no causal relationship between the two events.
       \vspace{1em}
    
    Based on this reasoning, the final answer is:
    \vspace{1em}
    
    \textbf{\{"Answer": "No"\}}
  \end{tcolorbox}

\end{case-box}
\end{figure*}

\begin{figure*}[htp]
\centering
\begin{case-box}[label={case:pattern_extractor}]{Extracting an Indirect Causal Chain}

  \begin{tcolorbox}[colback=gray!3, title=Input, fonttitle=\bfseries]
    Woman \textbf{[event2: Convicted]} of Double Murder In 2010 Kraft Foods \textbf{[event1: Rampage]} In NE Philadelphia.
  \end{tcolorbox}

  \begin{tcolorbox}[colback=gray!3, title=Pattern Extractor, fonttitle=\bfseries]
    \textbf{1.} The two events are:
    
       \hspace{1em} - EVENT X: Rampage
       
       \hspace{1em} - EVENT Y: Convicted
       \vspace{1em}
    
    \textbf{2.} Analyzing the direct causal relationship:
    
       \hspace{1em} - The text does not explicitly state that the "Rampage" directly caused the "Convicted" event. Instead, it indicates that a woman was convicted of double murder in connection with the rampage. Therefore, there is no direct causal relationship between the two events.
       \vspace{1em}
    
    \textbf{3.} Since there is no direct relationship, we need to analyze for indirect causal patterns. 
    \vspace{1em}
    
    \textbf{4.} Considering mediators:
    
       \hspace{1em} - The text mentions "double murder" as a significant event related to both the rampage and the conviction. The rampage likely led to the double murder, and the double murder is the basis for the conviction. Thus, "double murder" can be seen as a mediator (Z) between the two events.
       \vspace{1em}
    
    \textbf{5.} Analyzing the patterns:
    
       \hspace{1em} \textbf{- **Direct**:} Not applicable as established.
       
       \hspace{1em} \textbf{- **Coreference of X**:} Not applicable; there is no similar event to "Rampage" that has a causal relationship with "Convicted."
       
       \hspace{1em} \textbf{- **Coreference of Y**:} Not applicable; there is no similar event to "Convicted" that has a causal relationship with "Rampage."
       
       \hspace{1em} \textbf{- **Collider**:} Not applicable; while "double murder" is a mediator, it does not satisfy the collider condition since it does not act on both X and Y.
       
       \hspace{1em} \textbf{- **Fork**:} Not applicable; "double murder" does not act on both X and Y.
       
       \hspace{1em} \textbf{- **Chain**:} This pattern is satisfied because the rampage (X) leads to the double murder (Z), and the double murder (Z) leads to the conviction (Y). Thus, we have a causal chain: Rampage -> Double Murder -> Convicted.
       \vspace{1em}
    
    Final answer in JSON format:
    
    \textbf{\{"pattern": "Chain"\}}
  \end{tcolorbox}

\end{case-box}
\end{figure*}

\begin{figure*}[t]
\centering
\begin{case-box}[label={case:representative_positive}]{Representative Positive Example}

  \begin{tcolorbox}[colback=gray!3, title=Input, fonttitle=\bfseries]
    \textbf{[event1: Fire]} \textbf{[event2: destroys]} historic Mat Maid dairy building in Palmer.
  \end{tcolorbox}

  \begin{tcolorbox}[colback=gray!3, title=Input Events ConceptNet Path, fonttitle=\bfseries]
    "fire" is an antonym of "employ", and "employ" is related to "destroy", and "destroy" has the form of "destroys".
  \end{tcolorbox}

  \begin{tcolorbox}[colback=gray!3, title=Top Supporting Examples, fonttitle=\bfseries]
    \textbf{Example 1}
    
    \textbf{Text:} Wladimir Klitschko \textbf{[event2:won]} his 16th straight \textbf{[event1:fight]} today , successfully defending the WBA , WBO , and IBF heavyweight titles against Tony Thompson , stopping the 40 - year - old American in the sixth round of a fight that was never really close .
    
    \textbf{ConceptNet Path:} "fight" is the origin of "prizefight", and "prizefight" is derived from "prize", and "prize" receives the action of "won". (score = 0.426)
    
    \textbf{Tree similarity score:} 0.012
    
    \textbf{Pattern:} Direct
    \vspace{1em}
    
    \textbf{Example 2}
    
    \textbf{Text:} The woman who allegedly shot three co - workers — killing two– at the Kraft Plant Thursday night has been \textbf{[event2:charged]} with \textbf{[event1:murder]} , attempted murder , aggravated assault , burglary , and other related charges , says Philadelphia District Attorney Seth Williams .
    
    \textbf{ConceptNet Path:} "murder" has context in "legal", and "legal" is the context of "uncharged", and "uncharged" is derived from "charged". (score = 0.398)
    
    \textbf{Tree similarity score:} 0.009
    
    \textbf{Pattern:} Direct
  \end{tcolorbox}

  \begin{tcolorbox}[colback=gray!3, title=Generated Pattern at Inference, fonttitle=\bfseries]
    1. The two events in natural language are: 
    \hspace{1em} - EVENT X: A fire.
    \hspace{1em} - EVENT Y: Destroys a historic dairy building.
    \vspace{1em}
    
    2. The ConceptNet Path indicates that "fire" is related to "destroy" through the antonym of "employ." The Dependency Tree shows that "Fire" is the subject of the verb "destroys," which directly connects the two events.
    \vspace{1em}
    
    3. Analyzing for a direct causal relationship: The text states that a fire destroys a building. This indicates a direct causal relationship where the fire (X) directly causes the destruction (Y). Therefore, the causal pattern is "Direct."
    \vspace{1em}
    
    4. Since we have already established a direct causal relationship, we do not need to analyze for indirect causal patterns.
    \vspace{1em}
    
    Final answer in JSON format:
    \textbf{\{"pattern": "Direct"\}}
  \end{tcolorbox}

\end{case-box}
\end{figure*}

\begin{figure*}[t]
\centering
\begin{case-box}[label={case:error_analysis}]{Error Analysis Example}

  \begin{tcolorbox}[colback=gray!3, title=Input, fonttitle=\bfseries]
    After a quick survey of the scene , Owen said , McNutt's main \textbf{[event2:concern]} was keeping the \textbf{[event1:fire]} from spreading to the heating oil stored next door at the Crowley fuel distribution center .
  \end{tcolorbox}

  \begin{tcolorbox}[colback=gray!3, title=Input Events ConceptNet Path, fonttitle=\bfseries]
    "fire" is related to "danger", and "danger" is related to "concern".
  \end{tcolorbox}

  \begin{tcolorbox}[colback=gray!3, title=Top Supporting Examples, fonttitle=\bfseries]
    \textbf{Example 1}
    
    \textbf{Text:} Reports on Tuesday morning indicated that a fiber optic cable in the northern part of Egypt had been \textbf{[event1:damaged]} , resulting in the \textbf{[event2:loss]} of all telecommunications in Egypt's second largest city of Alexandria and parts of the Nile Delta region .
    
    \textbf{ConceptNet Path:} "damaged" is related to "waste", and "waste" is related to "loss". (score = 0.706)
    
    \textbf{Tree similarity score:} 0.004
    
    \textbf{Pattern:} Chain
    \vspace{1em}
    
    \textbf{Example 2}
    
    \textbf{Text:} Of the bombs that \textbf{[event1:landed]} inside the camp , one \textbf{[event2:detonated]} about a quarter mile from the NP compound .
    
    \textbf{ConceptNet Path:} "landed" is related to "property", and "property" is related to "mine", and "mine" is used for "detonate", and "detonate" has the form of "detonated". (score = 0.367)
    
    \textbf{Tree similarity score:} 0.039
    
    \textbf{Pattern:} Chain
  \end{tcolorbox}

  \begin{tcolorbox}[colback=gray!3, title=Generated Pattern at Inference, fonttitle=\bfseries]
    1. The two events are:
    \hspace{1em} - EVENT X: fire
    \hspace{1em} - EVENT Y: concern
    \vspace{1em}
    
    2. The ConceptNet Path indicates that "fire" is related to "danger," and "danger" is related to "concern." This suggests a conceptual link between the two events. The Dependency Tree shows that the concern is explicitly stated as McNutt's main concern regarding the fire, indicating a direct relationship.
    \vspace{1em}
    
    3. Analyzing for a direct causal relationship, the text does not explicitly state that the fire causes concern in a direct manner. Instead, it indicates that the concern is about preventing the fire from spreading. Therefore, the causal relationship is not direct.
    \vspace{1em}
    
    4. Since there is no direct causal relationship, we need to analyze for indirect causal patterns.
    \vspace{1em}
    
    5. The mediators in this context can be identified as "danger" (which is implied by the fire) and "spreading" (the concern is about preventing the fire from spreading). The fire (X) creates a situation of danger, which leads to concern (Y). Thus, we have:
    \hspace{1em} - X (fire) -> Z (danger)
    \hspace{1em} - Z (danger) -> Y (concern)
    \vspace{1em}
    
    6. Analyzing the patterns:
    \hspace{1em} - **Direct**: Not satisfied, as there is no explicit direct causal relationship.
    \hspace{1em} - **Coreference of X**: Not satisfied, as there is no similar event to "fire" that has a causal relationship with "concern."
    \hspace{1em} - **Coreference of Y**: Not satisfied, as there is no similar event to "concern" that has a causal relationship with "fire."
    \hspace{1em} - **Collider**: Not satisfied, as there are no mediators that both events relate to in a way that satisfies the collider structure.
    \hspace{1em} - **Fork**: Not satisfied, as there are no mediators that act on both events in a way that satisfies the fork structure.
    \hspace{1em} - **Chain**: Satisfied, as we have a mediator (danger) that connects fire to concern through the relationship of X -> Z -> Y.
    \vspace{1em}
    
    Final answer in JSON format:
    \textbf{\{"pattern": "Chain"\}}
  \end{tcolorbox}

\end{case-box}
\end{figure*}

We select four cases for illustration.

In Case \ref{case:cot_and_sere}, it can be seen that the CoT method exhibits ``causal hallucination'', whereas the \textbf{SERE} method correctly infers that there is no causal relationship in the given example.

Case \ref{case:pattern_extractor} demonstrates how the PatternExtractor in the \textbf{SERE} method accurately identifies a indirect causal relationship. Compared to Dr.ECI’s multi-agent approach, we accomplish this step using only a single LLM. By following our manually crafted rules, the LLM accurately extracts the causal pattern.

Case \ref{case:representative_positive} shows a representative positive example, including retrieved supporting examples, pattern induction, and final inference.

In Case \ref{case:error_analysis}, the model incorrectly predicts the patterns of the examples (the two examples should have been recognized as Direct; however, as we discuss in Appendix~\ref{appendix:causal_pattern}, GPT-4o-mini is less likely to predict Direct). Nevertheless, thanks to the assistance of the other two signals, the system is still able to identify the correct positive examples, which in turn guides the LLM to correctly complete the final reasoning.

\section{Causal Pattern}
\label{appendix:causal_pattern}

\begin{table*}[htp]
\begin{center}
\resizebox{\textwidth}{!}{
\scalebox{0.8}{
\begin{tabular}{c c c c}
    \toprule %
        \textbf{Causal Pattern} 
        & \textbf{Causal Graph} 
        & \textbf{Description} 
        & \textbf{Example} \\
    \midrule %
        \parbox[c]{2.5cm}{\centering Direct} 
        & \parbox[c]{2.5cm}{\centering \begin{tikzpicture}[scale=0.5]
            \node (X) [circle, draw, scale=0.9] {\scalebox{0.7}{$X$}};
            \node (Y) [circle, draw, right of=X, scale=0.9] {\scalebox{0.7}{$Y$}};
            
            \draw[->] (X) -- (Y);
        \end{tikzpicture}} 
        & \parbox[c]{5cm}{The text explicitly states a causal relationship between X and Y.} 
        & \parbox[c]{5cm}{\textellipsis to \textbf{shut down[X]} resulted in workers \textbf{losing[Y]} their jobs.} \\
    \midrule %
        \parbox[c]{2.5cm}{\centering Chain} 
        & \parbox[c]{2.5cm}{\centering \begin{tikzpicture}[scale=0.5]
            \node (X) [circle, draw, scale=0.9] {\scalebox{0.7}{$X$}};
            \node (Y) [circle, draw, right of=X, node distance=1.5cm, scale=0.9] {\scalebox{0.7}{$Y$}}; %
            \node (Z) [circle, draw, below of=X, xshift=0.75cm, scale=0.9, node distance=0.65cm] {\scalebox{0.7}{$Z$}}; %
            
            \draw[->] (X) -- (Z);
            \draw[->] (Z) -- (Y);
            \draw[dashed,->, bend left] (X) to (Y); %
        \end{tikzpicture}}
        & \parbox[c]{5cm}{There exist mediators (Z) that causally relate to both X and Y, with X acting on Z and Z acting on Y.}
        & \parbox[c]{5cm}{\textbf{Deforestation[X]} in \textellipsis soil \textbf{erosion[Z]}, which led to a \textbf{decline[Y]} in agricultural productivity.} \\
    \midrule %
        \parbox[c]{2.5cm}{\centering Collider} 
        & \parbox[c]{2.5cm}{\centering \begin{tikzpicture}[scale=0.5]
            \node (X) [circle, draw, scale=0.9] {\scalebox{0.7}{$X$}};
            \node (Y) [circle, draw, right of=X, node distance=1.5cm, scale=0.9] {\scalebox{0.7}{$Y$}}; %
            \node (Z) [circle, draw, below of=X, xshift=0.75cm, scale=0.9, node distance=0.65cm] {\scalebox{0.7}{$Z$}}; %
            
            \draw[->] (X) -- (Z);
            \draw[<-] (Z) -- (Y);
            \draw[dashed,->, bend left] (X) to (Y); %
        \end{tikzpicture}}
        & \parbox[c]{5cm}{There exist mediators (Z) that causally relates to both X and Y, with X acting on Z and Y acting on Z.}
        & \parbox[c]{5cm}{The \textbf{subsidy[X]} boosted electric vehicle \textellipsis  rising \textbf{demand[Y]} led \textellipsis a \textbf{strain[Z]} on battery \textellipsis .} \\
    \midrule %
        \parbox[c]{2.5cm}{\centering Fork} 
        & \parbox[c]{2.5cm}{\centering \begin{tikzpicture}[scale=0.5]
            \node (X) [circle, draw, scale=0.9] {\scalebox{0.7}{$X$}};
            \node (Y) [circle, draw, right of=X, node distance=1.5cm, scale=0.9] {\scalebox{0.7}{$Y$}}; %
            \node (Z) [circle, draw, below of=X, xshift=0.75cm, scale=0.9, node distance=0.65cm] {\scalebox{0.7}{$Z$}}; %
            
            \draw[<-] (X) -- (Z);
            \draw[->] (Z) -- (Y);
            \draw[dashed,->, bend left] (X) to (Y); %
        \end{tikzpicture}}
        & \parbox[c]{5cm}{There exist mediators (Z) that causally relate to both X and Y, with Z acting on X and Z acting on Y.} 
        & \parbox[c]{5cm}{\textellipsis economic \textbf{slowdown[Z]} led to a \textbf{decline[X]} in consumer spending and a \textbf{rise[Y]} in unemployment rates.} \\
    \midrule %
        \parbox[c]{2.5cm}{\centering Coreference}
        & \parbox[c]{2.5cm}{\centering \begin{tikzpicture}[scale=0.5]
            \node (X) [circle, draw, scale=0.9] {\scalebox{0.7}{$X$}};
            \node (X') [circle, draw, below of=X, scale=0.9] {\scalebox{0.7}{$X'$}};
            \node (Y) [circle, draw, right of=X, xshift=0.5cm, scale=0.9] {\scalebox{0.7}{$Y$}};
            \node (Y') [circle, draw, below of=Y, scale=0.9] {\scalebox{0.7}{$Y'$}};
            
            \draw[->, dashed, bend left] (X) to (Y);
            \draw[-] (X) -- (X');
            \draw[-] (Y) -- (Y');
            \draw[->] (X) -- (Y');
            \draw[->] (X') -- (Y);
            \draw[->, bend right] (X') to (Y');
        \end{tikzpicture}}
        & \parbox[c]{5cm}{There are different expressions of the same meaning for X and Y in the text, and these expressions are causally related.}
        & \parbox[c]{5cm}{\textellipsis the heavy rain, \textellipsis water level \textbf{rose[X]}, causing \textbf{inundation[Y]} in \textellipsis areas; the \textbf{increased[X']} rainfall then triggered a \textbf{flood[Y']} disaster.} \\
    \bottomrule %
\end{tabular}
}
}
\end{center}
\caption{All Causal Patterns used in this paper.}\label{tb:causal_pattern_full}
\end{table*}

The detailed introduction to the Causal Pattern is in Table~\ref{tb:causal_pattern_full}. 

The causal patterns used in this paper are adapted from Dr.ECI \cite{cai-etal-2025-dr}, but in order to emphasize ``structure'', we do not use \emph{Explicit Words}, \emph{Implicit Words}, and \emph{Causal Order} patterns. We use \emph{Direct} to represent explicit causal relationships, while the rest are classified as implicit causal relationships. Additionally, we require the LLM not to use common sense knowledge for pattern identification, as its rich pre-training data and powerful reasoning abilities might lead it to assume that many implicit causal relationships can be judged by common sense knowledge, thereby failing to extract a more accurate structure.

\section{Implementation of the Shortest Path algorithm}
\label{appendix:shortest_path_alg}

\begin{figure*}[htp]
\centering
\begin{minipage}{\textwidth}
\begin{lstlisting}[style=neo4j, caption={Cypher query for Shortest Path Algorithm}, label={code:shortest_path}]
MATCH p = shortestPath((a)-[r*0..{max_path_len}]-(b))
WHERE a.id = $start_id AND b.id = $end_id
RETURN p, [rel IN relationships(p) | {type: type(rel), start: startNode(rel).id, end: endNode(rel).id}] AS r
\end{lstlisting}
\end{minipage}
\end{figure*}

We directly perform Shortest Path matching on the Neo4j graph database using the Cypher query language (see Code~\ref{code:shortest_path}).

\section{Similarity Algorithm}
\label{appendix:similarity_alg}

\begin{table*}[!ht]
\centering
\resizebox{\textwidth}{!}{%
\scalebox{0.8}{%
\begin{tabular}{lll}
\toprule
\textbf{Relation} & \textbf{Forward Template} & \textbf{Inverse Template} \\
\midrule
HasContext & has context in & is the context of \\
MotivatedByGoal & is motivated by the goal of & motivates the goal of \\
FormOf & is a form of & has the form of \\
SimilarTo & is similar to & is similar to \\
HasA & has a & is owned by \\
dbpedia & is associated with the DBpedia concept of & has association from the DBpedia concept of \\
HasProperty & has the property of & is a property of \\
Causes & causes & is caused by \\
NotDesires & does not desire & is not desired by \\
HasPrerequisite & requires as a prerequisite & is a prerequisite for \\
PartOf & is part of & has as a part \\
Antonym & is an antonym of & is an antonym of \\
HasLastSubevent & has the last subevent of & is the last subevent of \\
MadeOf & is made of & is the material of \\
HasFirstSubevent & has the first subevent of & is the first subevent of \\
ReceivesAction & receives the action of & is the action performed on \\
RelatedTo & is related to & is related to \\
HasSubevent & has the subevent of & is a subevent of \\
DistinctFrom & is distinct from & is distinct from \\
InstanceOf & is an instance of & has an instance of \\
DerivedFrom & is derived from & is the origin of \\
UsedFor & is used for & uses \\
MannerOf & is a manner of & has as a manner \\
Desires & desires & is desired by \\
IsA & is a & has as a type \\
AtLocation & is located at & is the location of \\
CapableOf & is capable of & enables \\
EtymologicallyRelatedTo & is etymologically related to & is etymologically related to \\
Synonym & is a synonym of & is a synonym of \\
CreatedBy & is created by & creates \\
CausesDesire & causes the desire for & is desired because of \\
Entails & entails & is entailed by \\
DefinedAs & is defined as & defines \\
NotHasProperty & does not have the property of & is not a property of \\
\bottomrule
\end{tabular}
}%
}%
\caption{Mapping from predefined relations in Neo4j to natural language, including both forward and inverse forms.}
\label{tb:rel_nature_map}
\end{table*}

\begin{table*}[!ht]
\begin{center}
\resizebox{\textwidth}{!}{
\scalebox{0.8}{
\begin{tabular}{|c|c|c|}
    \hline %
        \textbf{Tree Node} 
        & \textbf{Description} 
        & \textbf{Weight}\\
    \hline %
        acl 
        & Noun clause modifier (structural, affects nominal structures) 
        & 2 \\
    \hline
        acomp 
        & Adjectival complement (supplementary description, does not affect core structure)   
        & 1 \\
    \hline
        advcl 
        & Adverbial clause modifier (enhances overall sentence modification)  
        & 3 \\
    \hline
        advmod 
        & Adverbial modifier (modifies verbs, moderate weight)  
        & 2 \\
    \hline
        agent 
        & Agent (critical, affects subject-verb structure)    
        & 5 \\
    \hline
        appos 
        & Appositive modifier (enhances sentence expression, affects structure)     
        & 3 \\
    \hline
        attr 
        & Attribute modifier (supplementary explanation, affects description)      
        & 1 \\
    \hline
        cc 
        & Coordinating conjunction (connects sentence structures, moderate weight)    
        & 2 \\
    \hline
        ccomp 
        & Complement clause (structural, complement clause has some impact)     
        & 3 \\
    \hline
        compound 
        & Compound modifier (enhances sentence structure expression, moderate weight)     
        & 3 \\
    \hline
        conj 
        & Coordinating word (moderate weight, used in coordinate structures) 
        & 2 \\
    \hline
        csubj 
        & Subject of a clause (structural, affects core sentence grammar) 
        & 5 \\
    \hline
        csubjpass 
        & Subject of a passive clause (structural, subject role within the clause)  
        & 5 \\
    \hline
        det 
        & Determiner (modifies nouns but does not change core structure) 
        & 1 \\
    \hline
        dobj 
        & Direct object (core component of the sentence, affects structure)
        & 4 \\
    \hline
        neg 
        & Negation modifier (negates sentence meaning, has some impact on structure)  
        & 2 \\
    \hline
        nounmod 
        & Noun modifier (modifies nouns, affects sentence structure)  
        & 2 \\
    \hline
        npmod 
        & Noun phrase as adverbial modifier (contributes to grammatical structure)  
        & 2 \\
    \hline
        nsubj 
        & Noun subject (one of the main components of the sentence, core element)   
        & 5 \\
    \hline
        nsubjpass 
        & Passive noun subject (core element, affects structure)    
        & 5 \\
    \hline
        nummod 
        & Numeral modifier (modifies numerals, minimal structural impact)  
        & 1 \\
    \hline
        oprd 
        & Object predicate (affects grammatical structure, part of the core)   
        & 3 \\
    \hline
        parataxis 
        & Parataxis (affects structure connection, moderate weight)   
        & 2 \\
    \hline
        pcomp 
        & Prepositional phrase complement (affects structure, but not the main clause core)    
        & 3 \\
    \hline
        pobj 
        & Prepositional object (important component of sentence structure)  
        & 4 \\
    \hline
        poss 
        & Possessive modifier (minimal impact, part of modification)  
        & 1 \\
    \hline
        preconj 
        & Preceding conjunction (connects sentences, affects structure)  
        & 1 \\
    \hline
        predet 
        & Preceding determiner (modifies nouns, minimal structural impact)  
        & 1 \\
    \hline
        prep 
        & Prepositional modifier (structural connection within the sentence, significant impact)  
        & 3 \\
    \hline
        prt 
        & Particle (minimal structural impact)  
        & 1 \\
    \hline
        quantmod 
        & Quantifier modifier (modifies quantifiers, minimal structural impact)  
        & 1 \\
    \hline
        relcl 
        & Relative clause modifier (has a significant impact on sentence structure) 
        & 3 \\
    \hline
        ROOT 
        & Sentence root node (core structure, most important)  
        & 5 \\
    \hline
        xcomp 
        & Open complement clause (affects sentence structure)  
        & 3 \\
    \hline
        
\end{tabular}
}
}
\end{center}
\caption{The dependency syntax tree nodes used in this paper and their corresponding weights. For nodes not listed in the table, we set their weight to 0.}\label{tb:tree_weights}
\end{table*}

\subsection{Path Serialization} 
We serialize a multi-hop path in the knowledge graph into a natural-language-like sequence by concatenating quoted node identifiers and direction-sensitive relation templates. 
For each relation $r$, a pair of textual templates $\langle t_{\rightarrow}, t_{\leftarrow}\rangle$ is provided to describe the forward and inverse directions (see Table~\ref{tb:rel_nature_map}). 
During serialization, we traverse the path sequentially; for each hop, the template is selected according to the direction of the edge in the path. 
For example, a path $a \xrightarrow{\text{HasA}} b \xleftarrow{\text{Causes}} c$ may be serialized as
\[
\texttt{"a" has a "b", and "b" is caused by "c".}
\]

This procedure produces a uniform sequence suitable for edit distance computation.

\subsection{Edit Distance based Path Similarity} 
After serializing each multi-hop path into a textual sequence, we measure the similarity between two paths based on the Levenshtein edit distance. 
Given two serialized paths $t_1$ and $t_2$, let $d(t_1, t_2)$ denote their edit distance. 
We normalize this distance by the length of the longer sequence to obtain a similarity score in the range $[0,1]$:
\[
\text{sim}(t_1, t_2) = 1 - \frac{d(t_1, t_2)}{\max(|t_1|, |t_2|)},
\]
where $|t_i|$ denotes the length of sequence $t_i$. 
A higher value of $\text{sim}(t_1, t_2)$ indicates greater structural resemblance between the two paths.

\subsection{Tree Edit Distance Based Text Similarity}
Let $T_1$ and $T_2$ denote the dependency trees of two text (for a single sentence, we directly parse its dependency tree; for a multi-sentence document, we first obtain the dependency tree of each sentence and then connect them to an artificial root node to construct the document-level dependency tree). We define a distance function $d(T_1, T_2)$ using tree edit distance, where the cost of updating, inserting, or deleting a node is determined by the dependency relation label. 

Specifically, for nodes with labels $l_1$ and $l_2$, the update cost is
\[
\text{update}(l_1, l_2) =
\begin{cases}
0, & l_1 = l_2,\\[1mm]
\text{cost}(l_1) + \text{cost}(l_2), & l_1 \neq l_2,
\end{cases}
\]

where the cost of each dependency label is weighted according to its structural importance (see Table~\ref{tb:tree_weights}). Core grammatical relations (\eg, \texttt{``nsubj''}, \texttt{``dobj''}, \texttt{``ROOT''}) receive higher weights, while modifiers receive lower weights. This weighted design ensures that changes to structurally central components contribute more to the edit distance than peripheral modifications.

Additionally, insertion and deletion costs are defined symmetrically:
\[
\text{update}(\emptyset, l) = \text{update}(l, \emptyset) = \text{cost}(l),
\]
where $\emptyset$ indicates missing label.

The normalized similarity score between two trees is then computed as
\[
\text{sim}(T_1, T_2) = \mathrm{e}^{-0.05 \cdot d(T_1, T_2)}.
\]

$\text{sim}(T1, T2)$ lies in the interval $(0, 1]$. We model similarity as an exponential function of the distance to ensure a controlled decay rate. A higher value of $\text{sim}(T_1, T_2)$ indicates greater structural resemblance.

\section{Prompts}
\label{appendix:prompts}

\begin{figure*}[htp]
\centering
\begin{minipage}{\textwidth}
\begin{prompt-box}[label={prompt:base}]{Base Method Prompt}
Given a text and two events (Event X and Event Y), please determine whether there is a causal relationship between Event X and Event Y (i.e., whether X -> Y or Y -> X).
\textbf{- Instructions:}
\hspace{1em}1. The final judgement you give must be either 'Yes' or 'No', and nothing else.
\hspace{1em}2. You need to organize the final answer in JSON format: \{"Answer": "Your answer, the answer must be either 'Yes' or 'No', and nothing else."\}.
\textbf{Text:} \{input\_text\};
\textbf{Event X:} \{source\};
\textbf{Event Y:} \{target\}.
Your answer in JSON format \{"Answer": "Your answer, the answer must be either 'Yes' or 'No', and nothing else."\}:
\end{prompt-box}
\end{minipage}
\end{figure*}

\begin{figure*}[htp]
\centering
\begin{minipage}{\textwidth}
\begin{prompt-box}[label={prompt:cot}]{CoT Method Prompt}
Given a text and two events (Event X and Event Y), please determine whether there is a causal relationship between Event X and Event Y.
\textbf{Text:} \{input\_text\};
\textbf{Event X:} \{source\};
\textbf{Event Y:} \{target\}.
Give step-by-step reasoning step and then give your answer in JSON format \{"Answer": "Your answer, the answer must be either 'Yes' or 'No', and nothing else."\}:
\end{prompt-box}
\end{minipage}
\end{figure*}

\begin{figure*}[htp]
\centering
\begin{minipage}{\textwidth}
\begin{prompt-box}[label={prompt:pattern_pos}]{Causal Pattern Extraction Prompt for Positive Sample}
\textbf{TEXT:} \{text\}, \textbf{EVENT X:} \{source\}, \textbf{EVENT Y:} \{target\}.
\textbf{QUESTION:} There is a causal relationship between EVENT X and EVENT Y. Please follow the following instructions to explain why there exists a causal relationship between X and Y based on the given text.
NOTE: You should put the final answer in JSON format: \{"pattern": THE CAUSAL PATTERN YOU FIND. DO NOT ANSWER "None" or "No".\}
\textbf{- Instructions:}
\hspace{1em}1. Analyze and determine whether X and Y have direct causal relationship, and meet the causal pattern rule "Direct". If so, answer causal pattern as "Direct"; If not, continue to analyze.
\hspace{1em}2. Determine which indirect causal pattern given below the given input and events satisfy. Note: If X and Y have the indirect causal relationship, they must satisfy to one of the following patterns.
\hspace{1em}3. Consider whether there are mediators between events X and Y: write down other events (or entities) that relates to X, and other events (or entities) that relates to Y, and determine whether there is any intersection between the events (or entities) that relate to both events. Note: Mediators can be given explicitly from the input text. If not given, you can also use common sense to think about whether there are implicit mediators.
\hspace{1em}4. Finally, analyze all the following patterns ONE-BY-ONE to determine whether the given text and events satisfy. DO NOT answer "None" or "No".
\textbf{- Pattern Rules:}
\hspace{1em} \textbf{Direct:} If the text explicitly states a causal relationship between X and Y without involving any mediating event (Z), then the causal connection is "Direct". This means that X directly influences Y, or Y directly influences X, with no intermediary mentioned.

\hspace{1em} \textbf{Coreference of X:} In the text, if an event with the same or similar meaning as the X can be found, and this similar event has a causal relationship with Y, then there is a causal relationship between X and Y.

\hspace{1em} \textbf{Coreference of Y:} In the text, if an event with the same or similar meaning as the Y can be found, and this similar event has a causal relationship with X, then there is a causal relationship between X and Y.

\hspace{1em} \textbf{Collider:} In the text, If there are one or multiple mediators (Z) that both events X and Y have causal relationship to: Then consider specific rule: First, it satisfies that it is related to X and Y respectively, then it satisfies that X acts on Z and Y acts on Z, i.e. X -> Z and Y -> Z. Therefore, it can be concluded that there is a causal relationship between X and Y.

\hspace{1em} \textbf{Fork:} In text, if there are one or multiple mediators (Z) that both events X and Y have causal relationship to: Then consider specific rule: First, it satisfies that it is related to X and Y respectively, then it satisfies that Z acts on X and Z acts on Y, i.e. Z -> X and Z -> Y. Therefore, it can be concluded that there is a causal relationship between X and Y.

\hspace{1em} \textbf{Chain:} In the text, if there are at least one or multiple mediators (Z) that both events X and Y have causal relationship to: Then consider specific rules: First, a mediator satisfies that it is related to X and Y respectively, then it satisfies that X acts on Z and Z acts on Y, i.e. they form a causal chain structure: X -> Z -> Y (or inversely, Y -> Z -> X). Then, it can be concluded that there is a causal relationship between X and Y.
NOTE: If one pattern rule is met, you DON'T need to analyze the remaining rules.
\textbf{EXAMPLE 1:}
\textit{TEXT}: The factory’s decision to shut down immediately resulted in hundreds of workers losing their jobs.
\textit{X}: shut down; \textit{Y}: losing
\textit{PATTERN}: Direct (The text clearly indicates a direct causal link between "factory’s decision to shut down" and "workers losing their jobs.")
\textbf{EXAMPLE 2:}
\textit{TEXT}: The company was accused of negligence in maintaining its pipelines, which were found to be leaking crude oil into the river. The oil spill caused significant harm to the local ecosystem.
\textit{X}: negligence in maintaining its pipelines; \textit{Y}: Harm
\textit{PATTERN}: Coreference ("The company was accused of negligence in maintaining its pipelines" and "pipelines leaking crude oil into the river" describe the same event in different ways. "Pipelines leaking crude oil" caused "harm to the ecosystem".)
\textbf{EXAMPLE 3:}
\textit{TEXT}: A major tech company introduced aggressive hiring policies, while a spike in tech startups also attracted talent to the industry. The resulting competition for skilled workers drove up average salaries in the tech sector.
\textit{X}: aggressive hiring policies; \textit{Y}: spike in tech startups; \textit{Z}: Competition for skilled workers.
\textit{PATTERN}: Collider ("Aggressive hiring policies" and "spike in tech startups" both increase competition for skilled workers (X -> Z, Y -> Z), which in turn drives up salaries, indirectly linking X and Y.)
\textbf{EXAMPLE 4:}
\textit{TEXT}: A global economic slowdown led to a decline in consumer spending and a rise in unemployment rates, as businesses struggled to stay profitable.
\textit{X}: decline in consumer spending; \textit{Y}: rise in unemployment rates
\textit{PATTERN}: Fork ("Global economic slowdown" causes both "a decline in consumer spending" (Z -> X) and "a rise in unemployment rates" (Z -> Y). This forms a Fork structure linking X and Y via Z.)
\textbf{EXAMPLE 5:}
\textit{TEXT}: Heavy deforestation in the region caused soil erosion, which eventually led to a decline in agricultural productivity.
\textit{X}: deforestation; \textit{Y}: decline; \textit{Z}: soil erosion
\textit{PATTERN}: Chain ("Heavy deforestation" leads to "soil erosion" (X -> Z), and "soil erosion" causes "a decline in agricultural productivity" (Z -> Y). This forms a causal chain)
\textbf{Your answer:}
\end{prompt-box}
\end{minipage}
\end{figure*}

\begin{figure*}[htp]
\centering
\begin{minipage}{\textwidth}
\begin{prompt-box}[label={prompt:pattern_inference}]{Causal Pattern Extraction Prompt for Inference Sample}
\textbf{TEXT:} \{text\}, \textbf{EVENT X:} \{source\}, \textbf{EVENT Y:} \{target\}.
\textbf{QUESTION:} Determine whether there is a causal relationship between EVENT X and EVENT Y.
NOTE: You should put the final answer in JSON format: \{"pattern": THE CAUSAL PATTERN YOU FIND. IF NO PATTERN RULES ARE MET, GIVE "No".\}
\textbf{- Instructions:}
\hspace{1em} 1. Analyze and determine whether X and Y have direct causal relationship, and meet the causal pattern rule "Direct". If so, answer causal pattern as "Direct"; If not, continue to analyze.
\hspace{1em} 2. Determine which indirect causal pattern given below the given input and events satisfy. Note: If X and Y have the indirect causal relationship, they must satisfy to one of the following patterns.
\hspace{1em} 3. Consider whether there are mediators between events X and Y: write down other events (or entities) that relates to X, and other events (or entities) that relates to Y, and determine whether there is any intersection between the events (or entities) that relate to both events. Note: Mediators can be given explicitly from the input text. If not given, you can also use common sense to think about whether there are implicit mediators.
\hspace{1em} 4. Finally, analyze all the following patterns ONE-BY-ONE to determine whether the given text and events satisfy. If no pattern rules are met, give "No"
\textbf{- Pattern Rules:}
\hspace{1em} \textbf{Direct:} If the text explicitly states a causal relationship between X and Y without involving any mediating event (Z), then the causal connection is "Direct". This means that X directly influences Y, or Y directly influences X, with no intermediary mentioned.

\hspace{1em} \textbf{Coreference of X:} In the text, if an event with the same or similar meaning as the X can be found, and this similar event has a causal relationship with Y, then there is a causal relationship between X and Y.

\hspace{1em} \textbf{Coreference of Y:} In the text, if an event with the same or similar meaning as the Y can be found, and this similar event has a causal relationship with X, then there is a causal relationship between X and Y.

\hspace{1em} \textbf{Collider:} In the text, If there are one or multiple mediators (Z) that both events X and Y have causal relationship to: Then consider specific rule: First, it satisfies that it is related to X and Y respectively, then it satisfies that X acts on Z and Y acts on Z, i.e. X -> Z and Y -> Z. Therefore, it can be concluded that there is a causal relationship between X and Y.

\hspace{1em} \textbf{Fork:} In text, if there are one or multiple mediators (Z) that both events X and Y have causal relationship to: Then consider specific rule: First, it satisfies that it is related to X and Y respectively, then it satisfies that Z acts on X and Z acts on Y, i.e. Z -> X and Z -> Y. Therefore, it can be concluded that there is a causal relationship between X and Y.

\hspace{1em} \textbf{Chain:} In the text, if there are at least one or multiple mediators (Z) that both events X and Y have causal relationship to: Then consider specific rules: First, a mediator satisfies that it is related to X and Y respectively, then it satisfies that X acts on Z and Z acts on Y, i.e. they form a causal chain structure: X -> Z -> Y (or inversely, Y -> Z -> X). Then, it can be concluded that there is a causal relationship between X and Y.
NOTE: If one pattern rule is met, you DON'T need to analyze the remaining rules.
\textbf{EXAMPLE 1:}
\textit{TEXT}: The factory’s decision to shut down immediately resulted in hundreds of workers losing their jobs.
\textit{X}: shut down; \textit{Y}: losing
\textit{PATTERN}: Direct (The text clearly indicates a direct causal link between "factory’s decision to shut down" and "workers losing their jobs.")
\textbf{EXAMPLE 2:}
\textit{TEXT}: The company was accused of negligence in maintaining its pipelines, which were found to be leaking crude oil into the river. The oil spill caused significant harm to the local ecosystem.
\textit{X}: negligence in maintaining its pipelines; \textit{Y}: Harm
\textit{PATTERN}: Coreference ("The company was accused of negligence in maintaining its pipelines" and "pipelines leaking crude oil into the river" describe the same event in different ways. "Pipelines leaking crude oil" caused "harm to the ecosystem".)
\textbf{EXAMPLE 3:}
\textit{TEXT}: A major tech company introduced aggressive hiring policies, while a spike in tech startups also attracted talent to the industry. The resulting competition for skilled workers drove up average salaries in the tech sector.
\textit{X}: aggressive hiring policies; \textit{Y}: spike in tech startups; \textit{Z}: Competition for skilled workers.
\textit{PATTERN}: Collider ("Aggressive hiring policies" and "spike in tech startups" both increase competition for skilled workers (X -> Z, Y -> Z), which in turn drives up salaries, indirectly linking X and Y.)
\textbf{EXAMPLE 4:}
\textit{TEXT}: A global economic slowdown led to a decline in consumer spending and a rise in unemployment rates, as businesses struggled to stay profitable.
\textit{X}: decline in consumer spending; \textit{Y}: rise in unemployment rates
\textit{PATTERN}: Fork ("Global economic slowdown" causes both "a decline in consumer spending" (Z -> X) and "a rise in unemployment rates" (Z -> Y). This forms a Fork structure linking X and Y via Z.)
\textbf{EXAMPLE 5:}
\textit{TEXT}: Heavy deforestation in the region caused soil erosion, which eventually led to a decline in agricultural productivity.
\textit{X}: deforestation; \textit{Y}: decline; \textit{Z}: soil erosion
\textit{PATTERN}: Chain ("Heavy deforestation" leads to "soil erosion" (X -> Z), and "soil erosion" causes "a decline in agricultural productivity" (Z -> Y). This forms a causal chain)
\textbf{Your answer:}
\end{prompt-box}
\end{minipage}
\end{figure*}

\begin{figure*}[htp]
\centering
\begin{minipage}{\textwidth}
\begin{lstlisting}[style=pythonstyle, caption={SERE final inference prompt}, label={code:sere_inference_prompt}]
def inference_by_examples_prompt(text: str, source: str, target: str, examples: list[dict[str, Any]]) -> str:
    instruction_prompt = '''Given a text, two events (Event X and Event Y). Based on the related examples, you need to determine whether there is a causal relationship between the given events X and Y. Please follow the instructions below and refer to the provided examples when answering.
###
Instructions:
You should refer to the examples but not be entirely influenced by them. Whether the events in the examples have a causal relationship DOES NOT affect whether the given events in the provided text have a causal relationship.
You should give step-by-step reasoning path before giving the final answer.
'''

    example_prompt = '''***Example {idx}***
Text: {text};
Event X: {source};
Event Y: {target};
Answer: {{"Answer": "{answer}"}}
'''

    examples_prompt = ''
    for idx, e in enumerate(examples, start=1):
        examples_prompt += example_prompt.format(idx=idx,
                                                 text=e['input_text'],
                                                 source=e['source'],
                                                 target=e['target'],
                                                 answer='Yes' if e['ground'] == 1 else 'No') + '\n'

    prompt = '''{instruction_prompt}
###
Here are some examples.
{examples_prompt}
###
Text: {text};
Event X: {source};
Event Y: {target};
Give step-by-step reasoning path, and then organize the final answer in JSON format: {{"Answer": "Your answer, the answer must be either 'Yes' or 'No', and nothing else."}}
Your response:
'''

    return prompt.format(instruction_prompt=instruction_prompt.format(text=text, source=source, target=target),
                         examples_prompt=examples_prompt,
                         text=text,
                         source=source,
                         target=target)
\end{lstlisting}
\end{minipage}
\end{figure*}

We present the prompts used in our paper, including the prompt for the Base method (Prompt~\ref{prompt:base}); the prompt for the CoT method (Prompt~\ref{prompt:cot}); the prompt for extracting causal patterns from positive samples in the training set (Prompt~\ref{prompt:pattern_pos}); the prompt for extracting causal patterns from inference samples (Prompt~\ref{prompt:pattern_inference}); and the prompt used for the final inference in the SERE method (Code~\ref{code:sere_inference_prompt}).

The difference between prompt~\ref{prompt:pattern_pos} and prompt~\ref{prompt:pattern_inference} is that the former one allows the LLM to know that the events in the sample must have a causal relationship, thus forcing it to infer the causal pattern and preventing it from answering "No". In these two prompts, we provide step-by-step reasoning instructions and definitions of different patterns. Additionally, we manually craft examples for each pattern to help the model extract them more accurately.

\end{document}